\documentclass{SCIS2025}

\begin{document}
\ArticleType{RESEARCH PAPER}
\Year{2025}
\Month{January}
\Vol{68}
\No{1}
\DOI{}
\ArtNo{}
\ReceiveDate{}
\ReviseDate{}
\AcceptDate{}
\OnlineDate{}
\AuthorMark{}
\AuthorCitation{}

\title{Robustness study of the bio-inspired musculoskeletal arm robot based on the data-driven iterative learning algorithm}{Title for citation}

\author[1]{Jianbo YUAN}{jianbo.yuan@sjtu.edu.cn}
\author[1]{Jing DAI}{}
\author[2]{Yerui FAN}{}
\author[2]{Yaxiong WU}{}
\author[1]{Yunpeng LIANG}{}
\author[1]{Weixin YAN}{xiaogu4524@sjtu.edu.cn}


\address[1]{School of Mechanical Engineering, Shanghai Jiao Tong University, Shanghai 200240, China}
\address[2]{School of Mechanical Engineering, University of Science and Technology Beijing, Beijing 100083, China}

\abstract{The human arm exhibits remarkable capabilities, including both explosive power and precision, which demonstrate dexterity, compliance, and robustness in unstructured environments. Developing robotic systems that emulate human-like operational characteristics through musculoskeletal structures has long been a research focus. In this study, we designed a novel lightweight tendon-driven musculoskeletal arm (LTDM-Arm), featuring a seven degree-of-freedom (DOF) skeletal joint system and a modularized artificial muscular system (MAMS) with 15 actuators. Additionally, we employed a Hilly-type muscle model and data-driven iterative learning control (DDILC) to learn and refine activation signals for repetitive tasks within a finite time frame. We validated the anti-interference capabilities of the musculoskeletal system through both simulations and experiments. The results show that the LTDM-Arm system can effectively achieve desired trajectory tracking tasks, even under load disturbances of 20\% in simulation and 15\% in experiments. This research lays the foundation for developing advanced robotic systems with human-like operational performance.}

\keywords{Bio-inspired robotics; musculoskeletal arm, data-driven iterative learning control, muscle model algorithm, robustness}

\maketitle

\section{Introduction}

Traditional robotic systems excel in high-precision and large-load operations, but achieving tasks that require robustness, dexterity, and flexibility necessitates high-precision sensors, high-precision structures, and advanced control algorithms. In situations where the absolute precision of sensing and control in each unit is not high, the human arm can effectively utilize its inherent structural characteristics, such as the serial and parallel hybrid kinematic structure and the rigid-flexible coupling dynamic characteristics, to achieve rapid, robust, safe, dexterous, and flexible operations through information processing in neural circuits~\cite{1,2,3}. Through the synergy of software and hardware, developing a neuromorphic intelligent robot system that embodies human-like structural characteristics and driving mechanisms holds significant inspirational and catalytic value for advancing novel high-performance robotic systems. However, simulating the musculoskeletal structure with physical devices poses significant challenges. Michael et al.~\cite{4} created the ‘Anthrob’ robot, which is a reduced version of the human upper limb with 13 compliant muscles and four joints, However, the complexity of muscle units makes the extension of multi-muscle actuation challenging. Kozuki et al.~\cite{5} designed the 'Kenshiro' shoulder-elbow joint using a parallel muscle device, which featured 21 muscle-driven units, a rib-surfaced thorax, muscle buffer devices, planar muscle devices, and open ball joints. The experiments successfully demonstrated abduction and elevation movements. However, the parallel muscle structure increases the complexity of structural coupling, making it challenging to accurately simulate real multi-muscle control. Kento et al.~\cite{6} developed a human-like forearm with a radioulnar joint to perform repetitive tasks with low operational accuracy. Subsequently, they designed the ‘MusashiLarm’ a fully functional musculoskeletal arm platform composed solely of joint modules, muscle modules, generic skeleton frames, muscle units, and a few attachments. Although modular and scalable design principles have been successfully implemented, uncertainties in muscle-to-bone contact and friction remain a significant challenge. The configuration of muscle layouts and the design of actuation pathways still require further optimization. Arne et al.~\cite{7} developed an arm-musculoskeletal robot powered by 10 pneumatic artificial muscles (PAMs). However, this design leads to issues such as low control frequency, poor accuracy, and redundant, complex attachments, which hinder its further application and development. Existing musculoskeletal robots are limited in simulating only localized joints or a few muscles, and lack the simulation of nonlinear muscle mechanical properties, resulting in insufficient maneuverability and operational capability. To address these gaps, this study conducted a comprehensive simulation of a full-morphology musculoskeletal arm, preserving the antagonistic muscle-driven mechanism, and designed a centralized artificial muscle architecture for a low-inertia robotic arm, paving the way for future research on humanoid musculoskeletal systems.

The abundance of redundant actuators in musculoskeletal robots presents substantial challenges in control. On one hand, the disparity between the control space and the task space dimensions increases the difficulty of optimization and solution. On the other hand, the structural characteristics of the rigid-flexible coupling and the nonlinear properties of muscle-like actuators make deriving an accurate controller difficult. Computational muscle control (CMC)~\cite{8} and antagonist inhibition control (AIC)~\cite{9} are the most widely utilized low-level control strategies for musculoskeletal robotics systems. Michael et al.~\cite{10} developed an adaptive neural network dynamic surface control (ANNDSC) to enhance trajectory tracking in the ‘Anthrob’ musculoskeletal robot. However, existing studies prioritize platform control yet neglect biological neuromuscular mechanisms. Model-based algorithms face challenges in generating precise activation signals due to muscle dynamics' high-dimensional nonlinearity and absent proprioceptive feedback. Jäntsch et al.~\cite{11} developed a control scheme for scalable joint-space actuation by decomposing the robot’s dynamics into hierarchical subsystems, resolving muscle-to-joint torque mapping in musculoskeletal systems. Kawaharazuka et al.~\cite{12}  developed a neural network model trained on joint-muscle nonlinear data to map time-series control inputs to task states, achieving real-time control. Buchler et al.~\cite{13} developed a Gaussian process-based probabilistic forward dynamics model to address the accurate control challenges in musculoskeletal robots. Driess et al.~\cite{14} integrated neural network-learned forward dynamics with sequential quadratic programming (SQP), enabling co-contraction modulation, stiffness adaptation, motion accuracy, and computational efficiency in musculoskeletal control. Existing control methods are effective for simple systems but struggle with highly redundant muscle-driven arms, nonlinear muscle models, and rigid-flexible coupling. Tendon-driven hysteresis and directional dependence further complicate model-based high-precision control. An iterative learning control (ILC) method based on dynamic linearization theory can improve performance in highly redundant and nonlinear robot systems~\cite{15,16,17}. It transforms discrete-time nonlinear systems into a virtual dynamic linearization model, using only I/O data for adaptive updates without the need for a system model. The muscle model’s inherent anti-interference capabilities~\cite{18,19}, combined with DDILC, can further enhance control performance and robustness.

In summary, a novel humanoid muscle-driven musculoskeletal robot was designed to improve disturbance rejection. To tackle the challenge of deriving explicit controllers for the complex nonlinear LTDM-Arm system, a DDILC controller was developed, and its convergence was demonstrated. Load disturbance experiments were then conducted to evaluate the disturbance rejection performance of the LTDM-Arm's structure and control algorithm.

\section{Mathematics Model of the LTDM-Arm}

This section introduces the musculoskeletal mathematical model. Additionally, it examines the relationship between muscle and joint space, as well as the kinematics and dynamics of each component within the LTDM-Arm system.

\subsection{Architecture of musculoskeletal model}

Muscles exhibit strong nonlinear properties and consist of muscle fibers and tendons. To describe activation kinetics, tension production, and muscle-tendon contraction, physiologists have studied isolated muscles ~\cite{20}. In biomechanical simulations, Hill-type muscle models are commonly used to predict both passive and active muscle forces ~\cite{21}. These models have been widely recognized for their completeness in representing muscle mechanisms and their computational efficiency ~\cite{20,22}. Therefore, Hill-type muscle models were chosen in this study to simulate the internal driving mechanism of the artificial muscle. The modeling accuracy of the biological muscle unit directly influences the precision with which the musculoskeletal model simulates motion. The Hill-type model consists of three components: the Contraction Element (CE), the Parallel Elastic Element (PEE), and the Series Elastic Element (SEE). This study employs the Hill-type muscle model, which is widely recognized for its biological accuracy, computational efficiency, and suitability for muscle control~\cite{23,24,25}. Subsequently, Thelen et al.~\cite{23} simplified the muscle-tendon unit into a composite model consisting of a nonlinear spring, an active contractile element, and a passive spring. The muscle model composes muscle tendon and muscle fiber connected in series through the pinnate angle ($\alpha$), as shown in Fig. 1. $\alpha=1$ is used in this experiment. Muscle activation a is reckoned by solving a non-linear first order differential equation, whose input is a dimensionless quantity between 0 and 1, called muscle excitation, denoted as $u$:
\begin{equation}
\dot{a} = \frac{u - a}{t_{a}(u, a)}
\end{equation}

\begin{equation}
t_{a}(u, a) =
\begin{cases}
t_{\text{act}}(0.5 + 1.5 a) & u \geq a \\
\frac{t_{\text{decact}}}{0.5 + 1.5 a} & u < a
\end{cases}
\end{equation}
where, $t_a(u, a)$ is a time constant that varies with the variation trend of muscle activation. $t_{\text{act}}$ and $t_{\text{decact}}$ are the activation time constant and the deactivation time constant, respectively, which are set to 10 ms and 40 $ms$~\cite{26}.

\begin{figure}[t]
	\centering
	\includegraphics[width=0.4\linewidth]{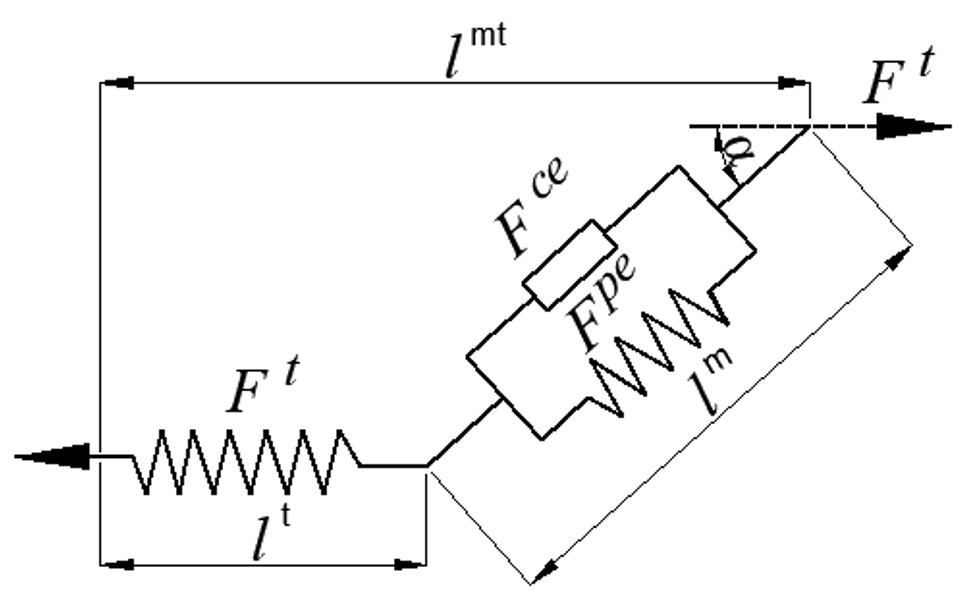}
	\caption{Hill-type muscle model description}
\end{figure}

Muscle tendons can only generate tension determined by the muscle properties. Furthermore, the muscle fiber force \( F_{\text{muscle}} \) is determined by the active contraction force \( F_{\text{active}} \) and the passive contraction force \( F_{\text{passive}} \). The active contraction force \( F_{\text{active}} \) is a function of the activation signal \( a \), the muscle length \( l \), and the muscle velocity \( v \). The passive contraction force \( F_{\text{passive}} \) only depends on the muscle length \( l \). The force of the muscle fiber \( F_{\text{muscle}} \) is defined as follows:
\begin{equation}
F^{t} - F^{m} \cos(\alpha) = 0
\end{equation}

\begin{equation}
F^{m} = F^{ce} + F^{pe} = a F_{0}^{m} f_{l}\left(\overline{l^{m}}\right) f_{v}\left(\overline{\dot{l}^{m}}\right) + f_{pe}\left(\overline{l^{m}}\right)
\end{equation}

\begin{equation}
\overline{\dot{l}^m} = f_{v}^{-1}\left(\frac{\frac{f^{t}\left(\overline{l^t}\right)}{\cos(a)}-f_{pe}\left(\overline{l^m}\right)}{af_{l}\left(\overline{l^m}\right)}\right)
\end{equation}

\begin{equation}
\overline{l^{mt}} = \overline{l^m}cos(\alpha) + \overline{l^t}
\end{equation}
where, \( f_{l} \) and \( f_{v} \) are functions of force-length and force-velocity, respectively; \( f_{pe} \) is the passive force of the muscle fiber; \( f_{v}^{-1} \) is the inverse function of the force-velocity; \( F^{t} \) is the tendon force.

The passive force-length relationship of muscle is represented by an exponential function:
\begin{equation}
f_{pe}(\overline{l^m}) = \frac{e^\frac{k^{pe}(\overline{l^m}-1)}{\epsilon_o^M}-1}{e^{k^{pe}}-1}
\end{equation}
where, \(k^{pe}\) is an exponential shape factor, and \(\epsilon_o^M\) is the passive muscle strain due to maximum isometric force.

The active force-length relationship of muscle is represented by a Gaussian function~\cite{27}:
\begin{equation}
f_l(\overline{l^m}) = e^\frac{-2(\overline{l^m}-1)^2}{\gamma}
\end{equation}
where, \(\gamma\) is a shape factor which is 0.45.

The force-velocity relationship is represented by an exponential function~\cite{28}:
\begin{equation}
f_v(\overline{\dot{l}^m}) = 1.6-1.6e^{[-\frac{1.1}{-(\overline{\dot{l}^m}+1)^4}+\frac{0.1}{(-\overline{\dot{l}^m}+1)^2}]}
\end{equation}

The force-strain relationship of tendon is represented by an exponential function during an initial nonlinear toe region and by a linear function thereafter:
\begin{equation}
f^t = 
\begin{cases}
\frac{F_{toe}^T}{e^{k_{toe^{-1}}}} \left( e^{\frac{k_{toe}\varepsilon^T}{\varepsilon_{toe}^T}-1} \right) & \varepsilon^T \leq \varepsilon_{toe}^T \\
k_{lin}(\varepsilon^T - \varepsilon_{toe}^T) + F_{toe}^T & \varepsilon^T > \varepsilon_{toe}^T
\end{cases}
\end{equation}
where, $\varepsilon^T$ is the tendon strain, $\varepsilon_{toe}^T$ is the tendon strain above which the tendon exhibits linear behavior, $k_{toe}$ is an exponential shape factor ($k_{toe}=3$) and $k_{lin}$ is a linear scale factor. The transition from nonlinear to linear behavior was prescribed to occur for tendon forces greater than $F_{toe}^T=0.33$. Requiring continuity of slopes at the transition resulted in $\varepsilon_{toe}^T=0.609\varepsilon_t^T$ and $k_{lin}=1.712\varepsilon_0^T$.

\subsection{Mathematical model of LTDM-Arm}

Due to the redundancy of joints in the musculoskeletal robot, for a given terminal movement velocity \(\dot{P}\) of the LTDM-Arm,, there are infinite solutions to the corresponding joint rotation velocity \(\dot{q}\). Solve in joint space by the following equation.
\begin{equation}
\dot{q}=J^+\dot{P}+(I-J^+J)K_q
\end{equation}
where, J is the Jacobian matrix of the joint space; \(J^+=J^T (JJ^T )^{-1}\) is moor-penrose inverse matrix of \(J\); \(K_q\) is a vector solution that represents an infinite number of possible solutions in the range of joint velocities.

The kinematic of the skeletons refers to the method of D-H Matrix. The Jacobian matrix \(L(q)\) maps the relationship between joint and muscle space. We can calculate muscle length \(l^{mt}\) according to following formula.
\begin{equation}
l^{mt} = L(q)q
\end{equation}
where $q$ is the joint angle.
Combined with the principle of virtual work~\cite{29}, we got the following relation:
\begin{equation}
dl = -L(q)dq
\end{equation}
\begin{equation}
\tau = -L(q) \cdot F^t + (I - L(q)^+L(q))K_m
\end{equation}
where, \(L(q)^+ = L(q)^T(L(q)L(q)^T)^{-1} \) is Moore-Penrose inverse matrix of \(L(q)\); \(K_m\) is a vector solution that represents an infinite number of possible solutions in the range of joint torque; \(q\) is a vector of desired joint variables, muscle length matrix \(dl = [dl_1, dl_2,...,dl_{15}]\) ; joint angle matrix \(dq=[dq_1, dq_2,...,dq_7]\) ; \(F^t = [f_1, f_2, f_3,...,f_{15}]\) is the muscle tension matrix;\(\tau = [\tau_1, \tau_2, \tau_3,...,\tau_7]\) is joints torque matrix. The minus sign represents the direction of muscle contraction.

In muscle mathematical models, muscles are treated as massless actuators, and in the muscle arm hardware platform, the cables at the anchor points are also considered as massless drivers driven by the cable-driven method. By integrating Equation (13) and muscle contraction dynamics Equation (3) into the muscle arm dynamic model, and obtain:
\begin{equation}
\begin{split}
H\left[-\dot{L}^+(q)\dot{l}^{mt}-L(q)\ddot{l}^{mt}\right]-\boldsymbol{\Theta}L^+(q)\dot{l}^{mt}-\mathbf{G}L^+(q)l^{mt} \\
=-\zeta\left(\left(aF^m f_l\left(\bar{l}^m\right)f_v\left(\overline{\dot{l}^m}\right)+f_{pe}\left(\overline{l^m}\right)\right)+\mathbf{J}^TF^{ext}+\boldsymbol{\tau}_f\right)
\end{split}
\end{equation}
where, \(F^m\)represents the maximum isometric force, which is related to the physical performance of the artificial muscle system and is an important indicator for evaluating the peak force of muscles. In an isometric state, the contraction velocity of muscle fibers \(\dot{l}^m=0\), and the maximum isometric force can be obtained when the activation \(a=1\).

\section{DDILC algorithm}

Convergent activation signal motion profiles are generated through repetition and control signal correction for feedforward or feedback control. During the LTDM-Arm's motion, variables such as muscle force, velocity, and position interact with each other. In this study, we propose a bias-format feedback control along the time axis, combined with feedforward control along the iteration axis. The controller design is as follows.
\subsection{Data-driven biased-format dynamic linearization method}

For the LTDM-Arm system (15), it is transformed into the following MIMO repetitive motion nonlinear non-affine discrete-time system~\cite{26}:
\begin{equation}
\boldsymbol{y}(t+1,k) = f(y(t,k), \boldsymbol{u}(t,k))
\end{equation}
where, \(u(t,k) \in [0,1]\) and \(\boldsymbol{y}(t,k) \in \mathbb{R}^m\)  denote the input activation signal control quantity at time $t$ and iteration \(k\), \(t=\{1,2,\ldots,T\}, k=1,2,\ldots\), positive integers for the termination time within finite duration, \(f(\cdot): \mathbb{R}^{2m} \mapsto \mathbb{R}^m\)for the corresponding variable dimension; for the unknown nonlinear vector function.

\textbf{Control Objective: }Design a learning controller to minimize the tracking error $\boldsymbol{e}(t+1, k)=\boldsymbol{y}_{\boldsymbol{d}}(t+1, k)-\boldsymbol{y}(t+1, k)=0$ at $k \rightarrow \infty$, where, \(\boldsymbol{y}_{\boldsymbol{d}}(t+1, k) \in \mathbb{R}^{m}\) is the desired output position vector of the system.

\textbf{Ideal Controller:} When only considering feedback error along the time axis, there exists an ideal learning feedback controller for system (16) such that $\boldsymbol{e}(t+1, k)=0$. Its mathematical description is as follows:
\begin{equation*}
\boldsymbol{u}^{b}(t, k)=\boldsymbol{C}\left(\boldsymbol{e}(t+1, k), \ldots, \boldsymbol{e}\left(t-n_{e, t}+2, k\right)\right) \tag{17}
\end{equation*}
where, $n_{e, t}$ is an unknown positive integer representing the size of the sliding window; $\boldsymbol{C}(\bullet): \mathbb{R}^{m n_{u, t}} \mapsto \mathbb{R}^{m}$ is an unknown nonlinear function, the considered tracking quantity is the tracking error within the sliding window at $\left[t-n_{e, t}+2, t+1\right]$, in the $k$-th iteration along the time axis.

\subsection*{3.2 Controller design}

The design of the following control objective is as follows:
\begin{equation*}
\boldsymbol{Q}=\frac{1}{2}\|\boldsymbol{e}(t+1, k)\|^{2}+\frac{1}{2} \lambda_{k}\left\|\Delta \boldsymbol{u}^{b}(t, k)\right\|^{2} \tag{18}
\end{equation*}
where, $\left\|\lambda_{k}\right\|>0$ is the weighting factor. The first term on the right-hand side represents control tracking performance, while the second term represents minimizing control input energy consumption.

Substituting Equation (17) into Equation (18) and differentiating with respect to the matrix variable $\boldsymbol{\Xi}_{e}(t, k)$, and get:

\begin{equation*}
\frac{\partial \boldsymbol{Q}}{\partial \boldsymbol{\Xi}_{e}(t, k)}=\boldsymbol{e}(t, k) \frac{\partial \boldsymbol{y}(t+1, k)}{\partial \boldsymbol{u}^{b}(t, k)} \Delta \boldsymbol{E}(t+1, k)+\lambda_{k} \boldsymbol{\Xi}_{e}(t, k) \Delta \boldsymbol{E}(t, k) \Delta \boldsymbol{E}^{T}(t, k) \tag{19}
\end{equation*}

Through gradient descent, the update algorithm for the learning control gains $\boldsymbol{\Xi}_{e}(t, k)$ is obtained as follows:
\begin{align*}
\hat{\boldsymbol{\Xi}}_{e}(t, k) & =\hat{\boldsymbol{\Xi}}_{e}(t-1, k)-\boldsymbol{\eta}_{k} \frac{\partial \boldsymbol{Q}}{\partial \hat{\boldsymbol{\Xi}}_{e}(t-1, k)} \\
& =\hat{\boldsymbol{\Xi}}_{e}(t-1, k)-\hat{\boldsymbol{\Xi}}_{e}(t-1, k) \boldsymbol{\eta}_{k} \lambda_{k} \Delta \boldsymbol{E}(t, k) \Delta \boldsymbol{E}^{T}(t, k)  \tag{20}\\
& +\boldsymbol{\eta}_{k} \frac{\partial \boldsymbol{y}(t+1, k)}{\partial \boldsymbol{u}^{b}(t, k)} \boldsymbol{e}(t, k) \Delta \boldsymbol{E}^{T}(t, k)
\end{align*}
where, $\hat{\boldsymbol{\Xi}}_{e}(t, k)$ is estimated value of $\boldsymbol{\Xi}_{e}(t, k)$.\\

Because system (15) exhibits strong nonlinearity, $\boldsymbol{e}(t+1, k)$ and the partial derivatives $\partial y(t+1, k) / \partial u^{b}(t, k)$ cannot be directly obtained, conventional algorithms typically require training on a large amount of offline data. In this section, we adopt a relatively simple along-the-time-axis tight format dynamic linearization method~\cite{30} to obtain an equivalent data model, given by the formula:
\begin{equation*}
\Delta \boldsymbol{y}(t+1, k)=\boldsymbol{\Phi}(t, k) \Delta u^{b}(t, k) \tag{21}
\end{equation*}
where, $\Delta \boldsymbol{y}(t+1, k)=\boldsymbol{y}(t+1, k)-\boldsymbol{y}(t, k)$ is unknown matrix.

$$
\boldsymbol{\Phi}(t, k)=\left[\begin{array}{cccc}
\Phi_{11}(t, k) & \Phi_{12}(t, k) & \cdots & \Phi_{1 m}(t, k) \\
\Phi_{21}(t, k) & \Phi_{22}(t, k) & \cdots & \Phi_{2 m}(t, k) \\
\vdots & \vdots & \ddots & \vdots \\
\Phi_{m 1}(t, k) & \Phi_{m 2}(t, k) & \cdots & \Phi_{m m}(t, k)
\end{array}\right] \in \mathbb{R}^{m \times m}
$$

The $\boldsymbol{\Phi}(t, k)$ is partitioned Jacobian matrix (PJM) of the nonlinear system (16), satisfies $\|\boldsymbol{\Phi}(t, k)\| \leq b_{2}$, where $b_{2}>0$ is unknown constants.

\textbf{Assumption 1:} $\left|\boldsymbol{\Phi}_{i, j}(t, k)\right| \leq c_{1}, \quad\left|\boldsymbol{\Phi}_{i, i}(t, k)\right| \leq a c_{2}, \quad i=1, \ldots, m, j=1, \ldots, m, \quad i \neq j, \quad a \geq 1$, $c_{2}>c_{1}(2 a+1)(m-1)$, and element symbols remain unchanged.

Through offline data prediction methods, constructing the input-output mapping of a system is an approximation. By combining the virtual dynamic linearization model (20) and the updating algorithm (19), the process of matrix inversion is avoided, and estimation is performed according to the improved projection algorithm ~\cite{30}. The value of $\boldsymbol{\Phi}(t, k)$ is calculated using the following formula:
\begin{equation*}
\hat{\boldsymbol{\Phi}}(t, k)=\hat{\boldsymbol{\Phi}}(t-1, k)+\frac{\rho_{t}\left(\Delta \boldsymbol{y}(t, k)-\hat{\Phi}(t-1, k) \Delta \boldsymbol{u}^{b}(t-1, k)\right)\left(\Delta \boldsymbol{u}^{b}(t-1, k)\right)^{T}}{\mu_{t}+\left\|\Delta \boldsymbol{u}^{b}(t-1, k)\right\|^{2}} \tag{22}
\end{equation*}
where, $\rho_{t} \in(0,1]$ is the step size factor of $\Phi(t, k)$ at time $t$, and $\mu_{t}>0$ is the weight factor. $\hat{\Phi}(t, k)$ is estimation of $\boldsymbol{\Phi}(t, k)$. When any of the following conditions occur: $\left\|\hat{\boldsymbol{\Phi}}_{i, j}(t, k)\right\|<c_{2},\left\|\hat{\boldsymbol{\Phi}}_{i, j}(t, k)\right\|>a c_{2}$, $\left\|\hat{\boldsymbol{\Phi}}_{i, j}(t, k)\right\|<c_{1},\left\|\hat{\boldsymbol{\Phi}}_{i, i}(t, k)\right\|>\operatorname{ac}_1, \operatorname{sign}\left(\hat{\boldsymbol{\Phi}}_{i, j}(t, k)\right) \neq \operatorname{sign}\left(\hat{\boldsymbol{\Phi}}_{i, j}(1, k)\right)$, obtain $\hat{\boldsymbol{\Phi}}_{i, j}(t, k)=\hat{\boldsymbol{\Phi}}_{i, j}(1, k), i \neq j$.

According to formulas (21) and (22), we can obtain the estimation methods for $\boldsymbol{e}(t+1, k)$ and the partial derivative term $\partial y(t+1, k) / \partial u^{b}(t, k)$ :

\begin{gather*}
\hat{\boldsymbol{e}}(t+1, k)=\boldsymbol{y}_{d}(t+1, k)-\boldsymbol{y}(t, k)-\hat{\boldsymbol{\Phi}}(t, k) \Delta \mathbf{u}^{b}(t, k)  \tag{23}\\
\frac{\partial \hat{\boldsymbol{y}}(t+1, k)}{\partial \mathbf{u}^{b}(t, k)}=\hat{\boldsymbol{\Phi}}(t, k) \tag{24}
\end{gather*}

Building upon dynamic linearization feedback control, adding a feedforward term results in the following control law:

\begin{equation*}
\boldsymbol{u}(t, k)=\boldsymbol{u}^{b}(t, k)+\boldsymbol{u}^{f}(t, k) \tag{25}
\end{equation*}
where, $\boldsymbol{u}^{f}(t, k)=\boldsymbol{u}^{f}(t, k-1)+\boldsymbol{\beta e}(t+1, k-1), \boldsymbol{\beta}$ is gain coefficient matrix, and satisfies the condition of $\left\|1-\boldsymbol{\beta} \boldsymbol{f}_{u}\left(\boldsymbol{\varphi}(t+1, k), \ldots, \varphi\left(t-l_{e, t}+2, k\right)\right)\right\|<1$. Its constraint conditions are provided in the convergence proof.

The learning control law (25), feedback control update algorithm (20) (22), and estimation algorithms (23) (24) constitute the data-driven DDILC method of this chapter. Specific details are provided in Table 1.

\begin{table}[t]
\centering
\caption{Data-Driven Feedforward-Feedback DDILC Computation Process Without Model}
\begin{tabular}{p{0.12\textwidth}p{0.85\textwidth}}
\hline
\multicolumn{2}{c}{The computation process} \\
\hline
Step 1 & Set $t = 1$, initialize the controlled system (16) to generate input-output data and estimation values $\hat{\boldsymbol{\Phi}}(t,1)$, and randomly initialize $\hat{\boldsymbol{\Xi}}_e(t,1)$ \\

\multirow{2}{*}{Step 2} & Calculate the control input vector according to the following learning control law: \\
 & $\boldsymbol{u}(t,k) = \boldsymbol{u}^b(t,k) + \hat{\boldsymbol{\Xi}}_e(t,k)\boldsymbol{\Delta E}(t+1,k) + \boldsymbol{u}^f(t,k-1) + \beta\boldsymbol{e}(t+1,k-1)$ \hfill (26) \\
 & Follow the reset mechanism. $\boldsymbol{u}(t,k) = \text{sat}(\underline{\boldsymbol{u}}(t), \overline{\boldsymbol{u}}(t))$, Measured system output. $\boldsymbol{y}(t,k)$ \\

Step 3 & Estimate $\hat{\boldsymbol{\Phi}}(t-1,k-1)$ using formula (22) and follow the reset mechanism $\hat{\boldsymbol{\Phi}}(t,k) = \text{sat}(\underline{\boldsymbol{\Phi}}(t), \overline{\boldsymbol{\Phi}}(t))$ \\

\multirow{2}{*}{Step 4} & Calculate the Partial Partial Jacobian Matrix (PPJM) $\boldsymbol{\Xi}(t,k)$ using formula (24) and follow the reset mechanism \\
 & $\hat{\boldsymbol{\Xi}}(t,k) = \text{sat}(\underline{\boldsymbol{\Xi}}(t), \overline{\boldsymbol{\Xi}}(t))$ \\

Step 5 & Let $t = t+1$ and loop back to (2) \\
\hline
\multicolumn{2}{p{0.97\textwidth}}{Note: $\text{sat}(b,\bar{b})$ denotes the projection of variable onto the interval $[\underline{b},\bar{b}]$, $\underline{b}, \bar{b} \in \mathbb{R}$.} \\
\end{tabular}
\end{table}

\begin{figure}[t!]
	\centering
	\includegraphics[width=1\linewidth]{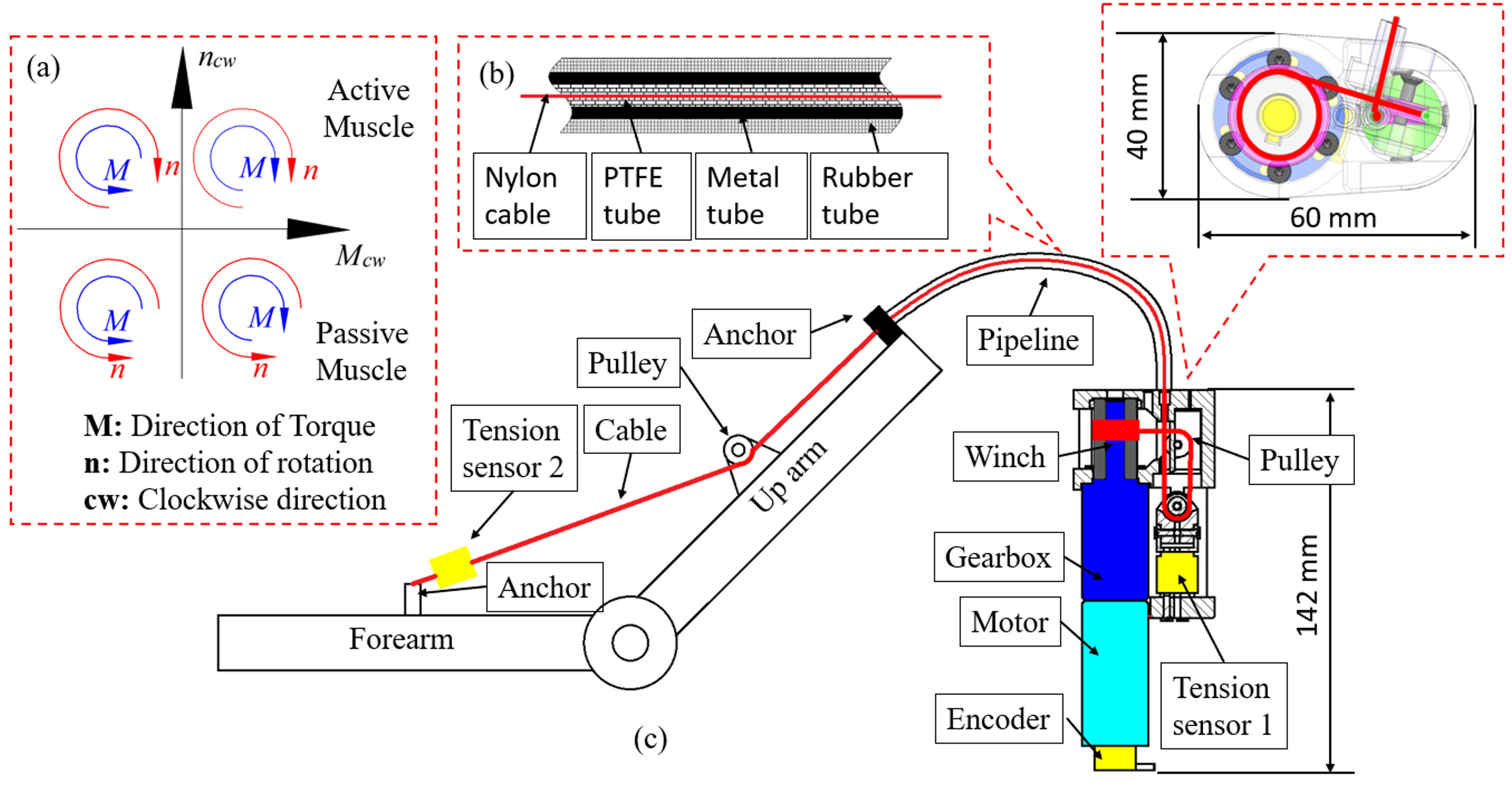}
	\caption{Modularized artificial muscular system (MAMS) scheme: (a) Quadrant diagram of DC motor motion states; (b) Schematic diagram of tube-rope transmission; (c) Schematic diagram of single-joint muscle-driven principle.}
\end{figure}

\section*{4 Design of the LTDM-Arm}

Muscle tissue consists of tendons and fibers, which can either generate tension to drive skeletal motion or relax under external forces to exhibit passive compliance. The challenges of simulating artificial muscles-such as transmission in confined spaces, high energy density, and passive compliance-make physical device replication difficult. A MAMS (Motor-Cable Artificial Muscle System) has been designed, which includes a motor-cable artificial muscle and a tendon-sheath-pulley system, as shown in Fig.2. A DC motor has been selected for power due to its high energy density, mature control technology, and convenient accessories. A force sensor is installed at the impact point to provide real-time rope tension feedback. The MAMS has dimensions of 142 mm (length), 60 mm (width), and 40 mm (height). The winch is driven by a motor and reducer to adjust the cable tension. The bracket is fabricated using high-precision 3D printing with Nylon (HP3DHR-PA12). To prevent wear and breakage of the tungsten rope under large curvature transmission, a high-strength polyester fiber cable has been chosen. The cable sheath consists of lubricated polytetrafluoroethylene (PTFE), a metal tube, and a rubber tube. A 0.2 mm gap separates the inner tube and cable, allowing free movement of the cable and ensuring effective lubrication. This MAMS design is intended for use in musculoskeletal robots, with additional details provided in previous work~\cite{31}.

It is known that the human arm contains more than 50 muscles, with multiple muscles collaborating to drive a single joint, allowing for controlled variable joint stiffness~\cite{23}. This specialized structure enables the arm to perform both fine operations and explosive movements, demonstrating flexibility and robustness. However, constructing a musculoskeletal robot resembling the human arm is challenging due to its structural complexity. Based on the designed MAMS and muscle configuration~\cite{32}, the LTDM-Arm platform, which closely mimics the human arm's anatomy, has been developed, as shown in Fig.3. The LTDM-Arm consists of seven degrees of freedom (DOF) (three DOF at the shoulder, one at the elbow, one at the forearm, and two at the wrist), four main skeleton components (shoulder blade, humerus, ulna, and radius), and 15 MAMSs (seven muscles at the shoulder, two at the elbow, two at the forearm, and four at the wrist). The upper arm, forearm, and hand lengths are $380 \mathrm{~mm}, 340 \mathrm{~mm}$, and 262 mm , respectively. The joint range of motion is [0\~{}0.4, $0 \sim 1.1,0 \sim 1.1,0 \sim 1.57,0 \sim 1.57,-1 \sim 1,-1 \sim 1]$ radians. A high-precision sensing system is required, including individual tension sensors for each MAMS and angle sensors for each joint. Each DOF is controlled by at least two muscles, with each muscle governing only one joint, which reduces coupling between the muscle and joint spaces, assuming antagonistic muscle connections are met. This design not only simulates the anatomical structure of the human arm but also adheres to mechanical design principles. It offers advantages such as a lightweight robotic arm, a flexible number of muscles, and reconfigurable anchor points, providing a solid experimental foundation for the control mechanisms and benefits of musculoskeletal systems.

\begin{figure}[t]
	\centering
	\includegraphics[width=1\linewidth]{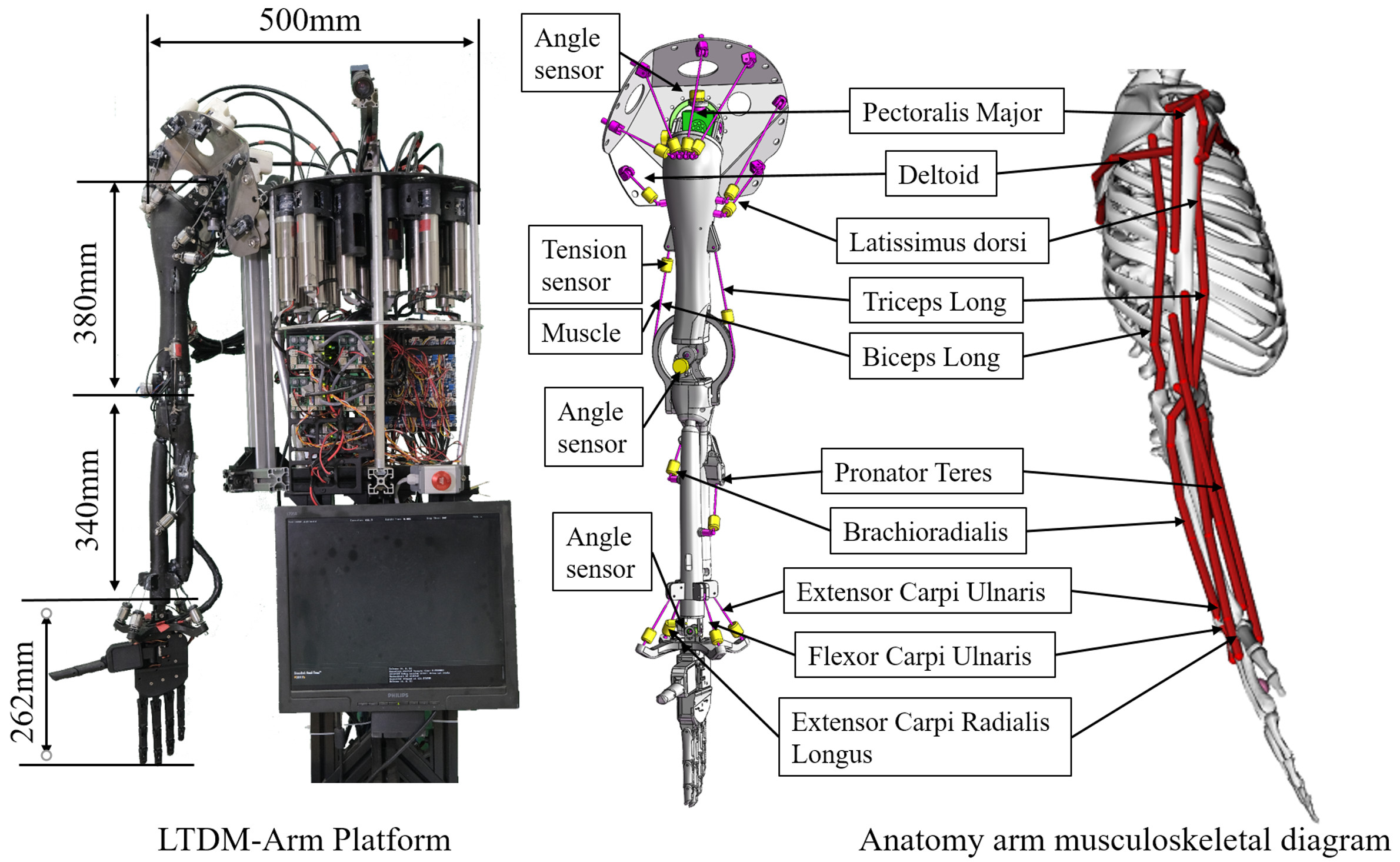}
	\caption{(Color online) LTDM-Arm design scheme.}
\end{figure}

\begin{figure}[t!]
	\centering
	\includegraphics[width=1\linewidth]{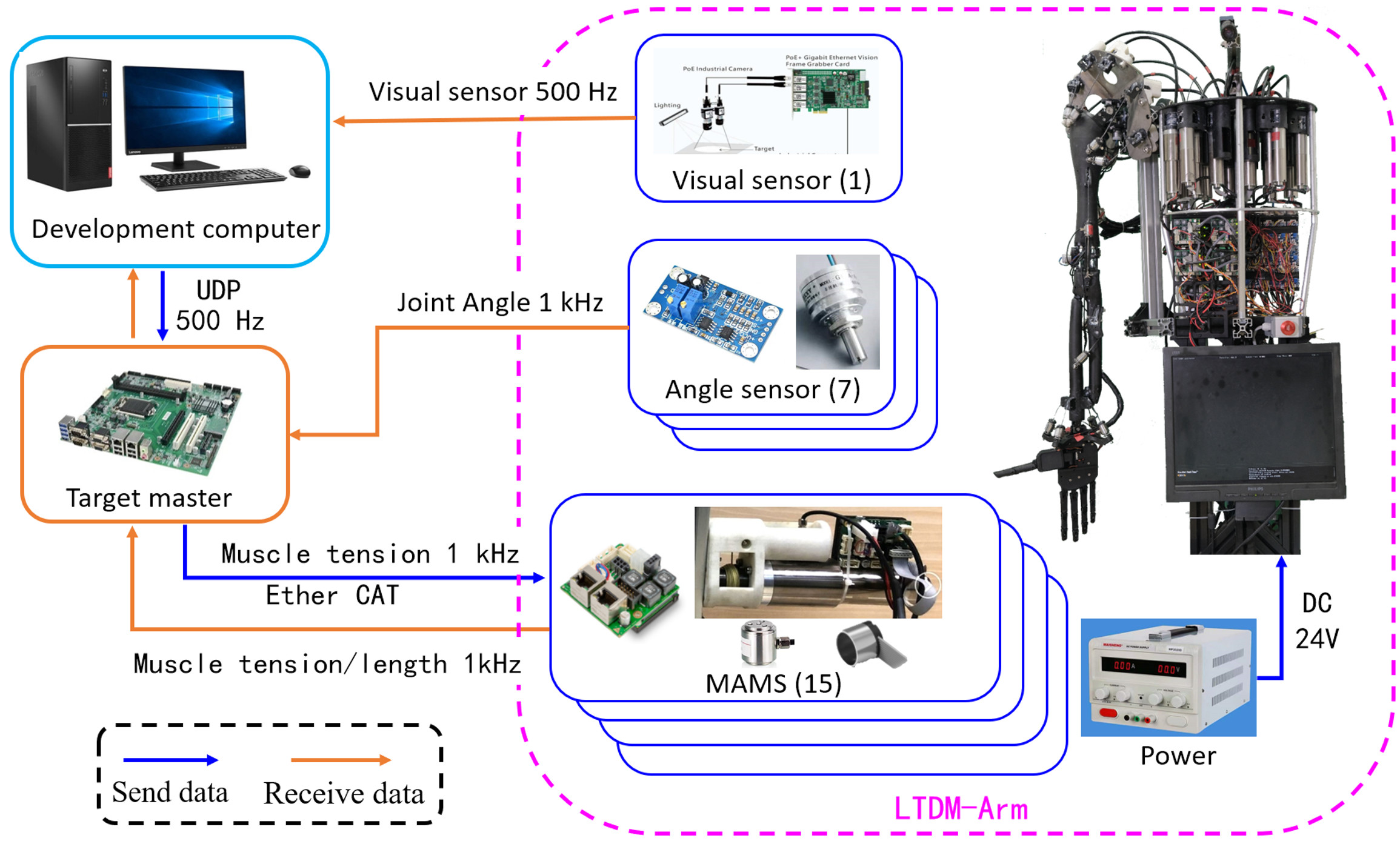}
	\caption{The circuit architecture of LTDM-Arm}
\end{figure}

\begin{figure}[t!]
	\centering
	\includegraphics[width=0.5\linewidth]{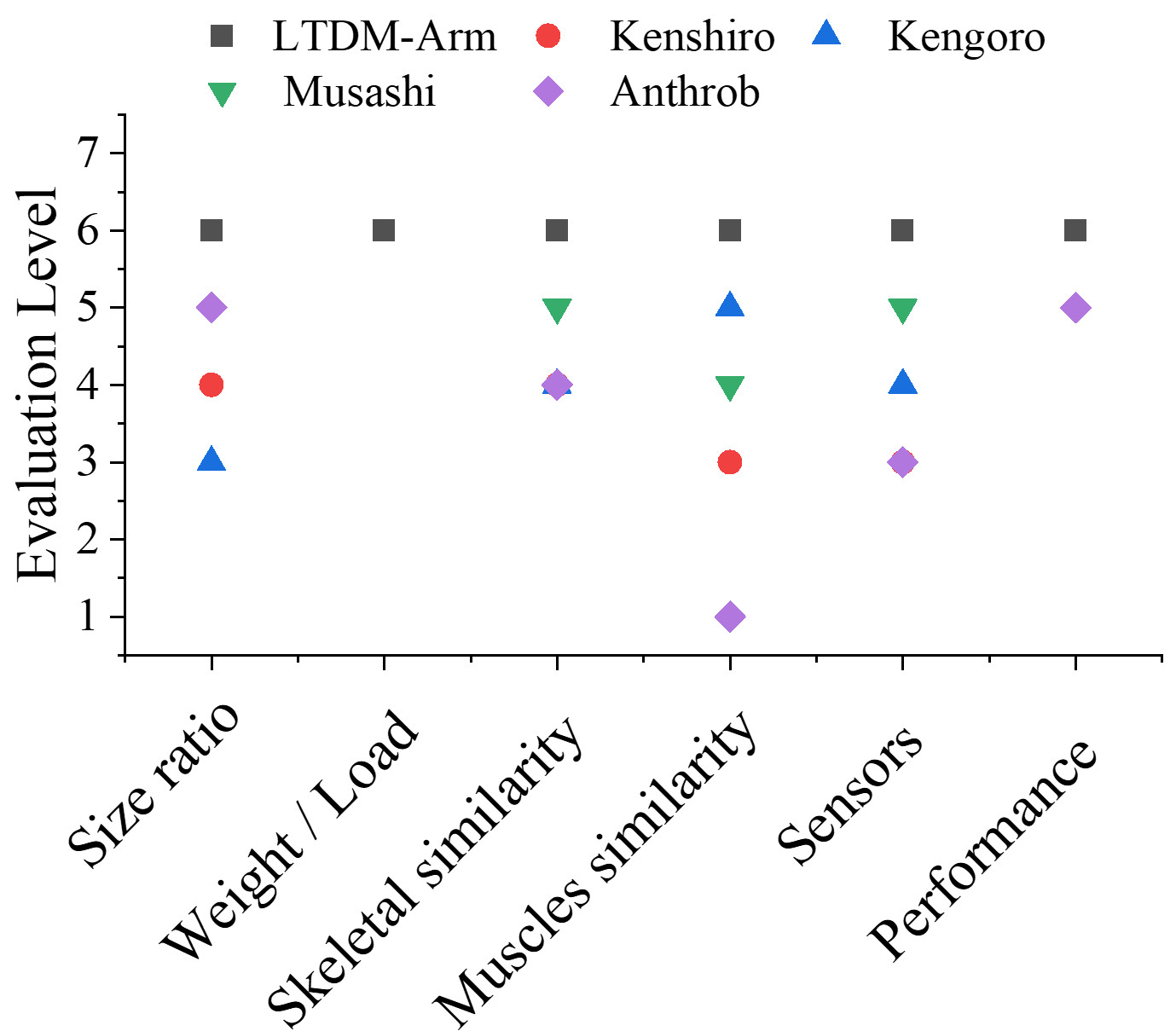}
	\caption{Performance comparison results of LTDM-Arm with other similar devices}
\end{figure}

The LTDM-Arm circuit architecture consists of three stages: the higher-level (algorithmic control), lower-level (protocol conversion and data processing hub), and physical-level (motors, encoders, sensors). The high-level control communicates with the lower-level and visual sensors via the UDP protocol, facilitating the transmission of control commands and feedback from the motors and sensors. The lower-level and physical layers use the Ether CAT protocol for information exchange, with each MAMS, joint angle sensor, and muscle tension sensor acting as communication nodes. These nodes transmit data to the control master station at a frequency of up to 1000 Hz . The circuit implementation is illustrated in Fig.4.

Based on data from an adult male arm ~\cite{33}, a comparative analysis was performed on representative musculoskeletal arm prototypes developed both domestically and internationally. The platforms compared include the 'Kenshio' ~\cite{34}, 'Kengoro' ~\cite{35}, and 'Musashi' ~\cite{36} musculoskeletal arm sections developed by the Tokyo Institute of Technology in Japan, as well as the 'Anthrob' musculoskeletal arm ~\cite{4} developed by the Technical University of Munich in Germany. Using the comparison criteria outlined in ~\cite{37}, the analysis focused on six main aspects: 1) Size ratio; 2) Weight/Load; 3) Skeletal similarity; 4) Muscle similarity; 5) Sensors; and 6) Performance, as shown in Fig.5. Based on the comparative results, the LTDM-Arm designed in this study demonstrates advanced configuration and performance, with a higher degree of human-likeness. This provides a valuable reference for research on novel humanoid musculoskeletal robotic systems.

\begin{figure}[t]
    \centering
    \subfloat[Numerical simulation trajectory]{
        \includegraphics[width=0.5\linewidth]{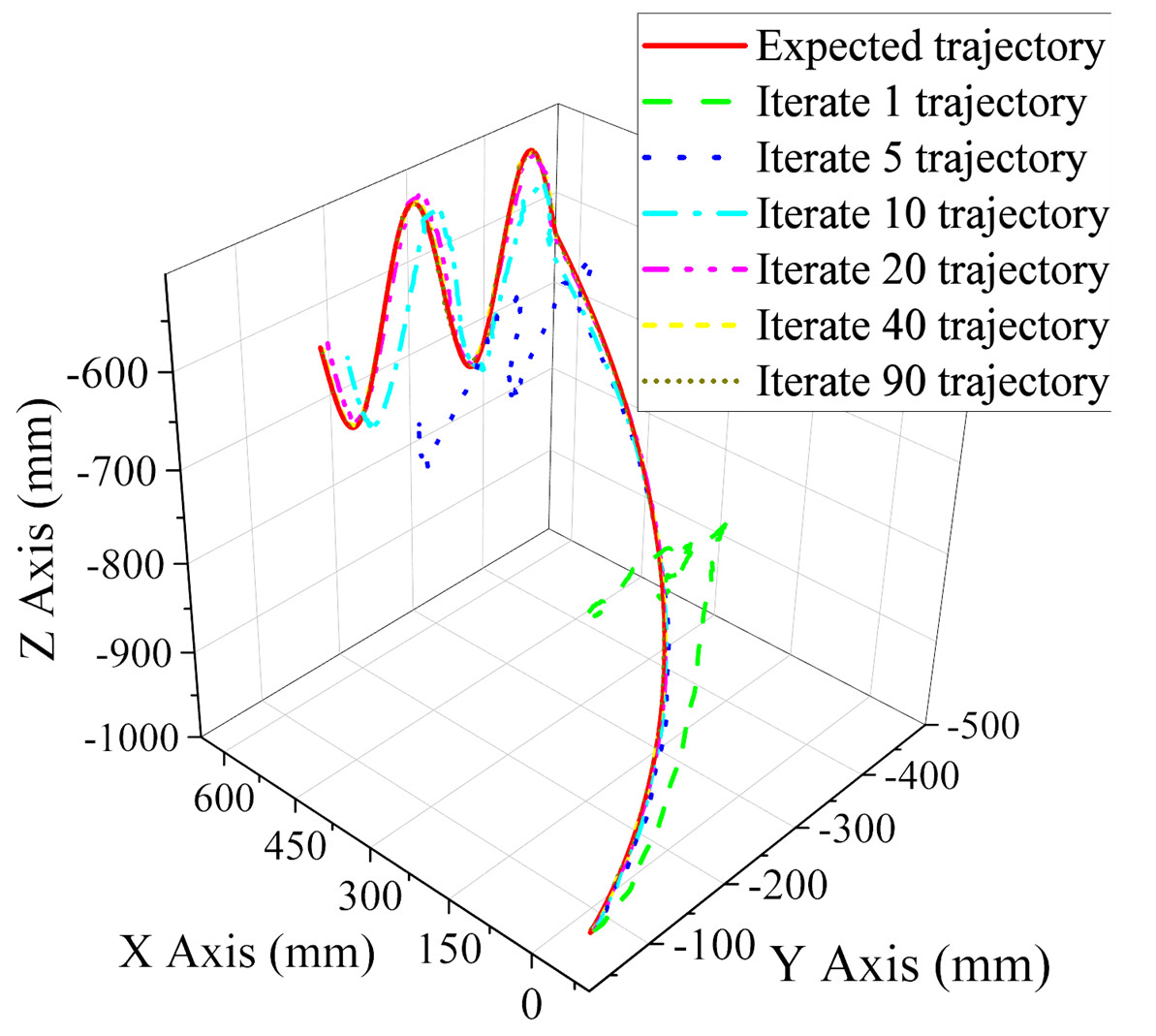}
    } \hfill
    \subfloat[Numerical simulation trajectory error]{
        \includegraphics[width=0.45\linewidth]{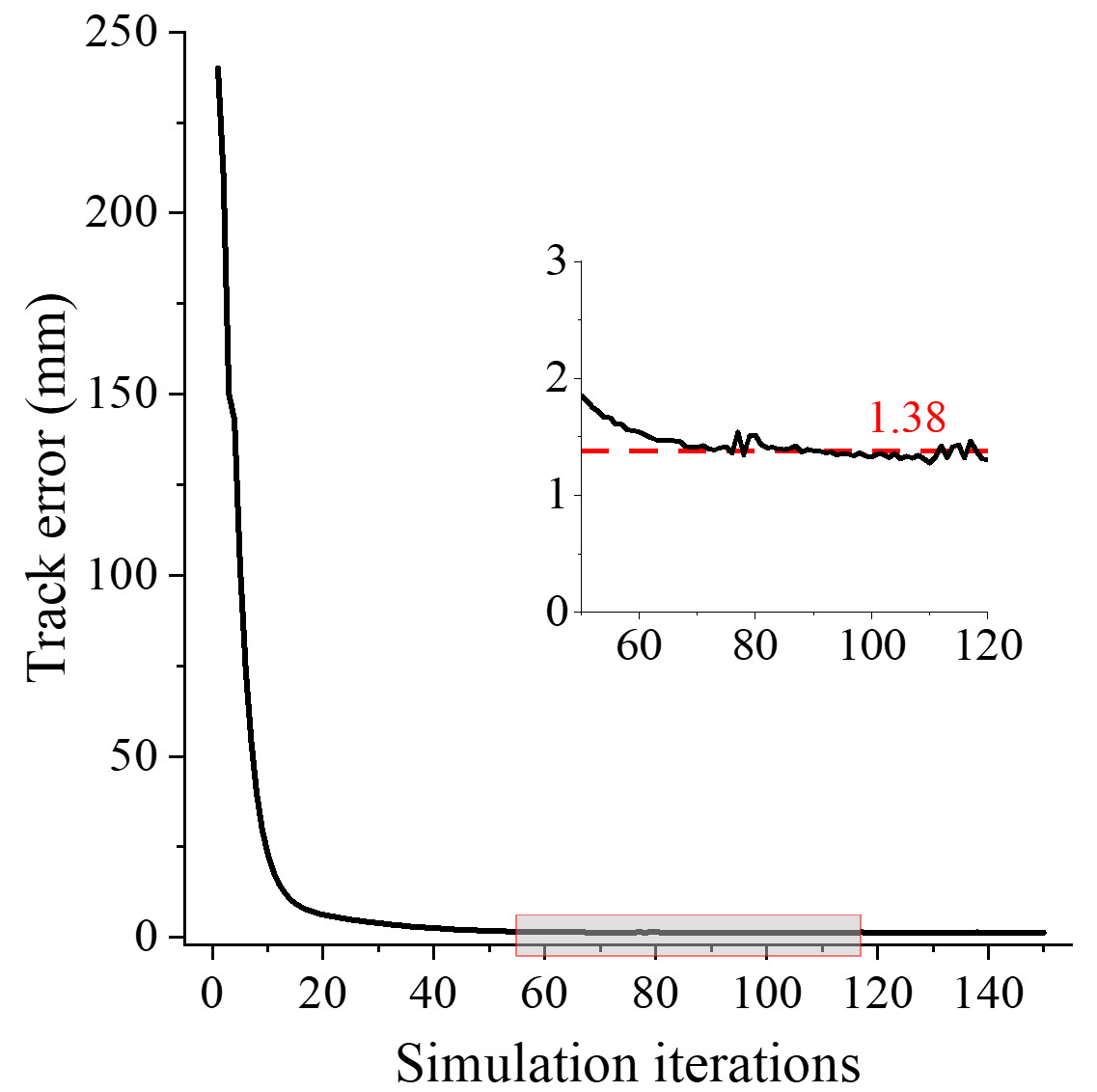}
    }
    \caption{Trajectory and error results of DDILC numerical simulation}
\end{figure}

\begin{figure}[t!]
    \centering
    \includegraphics[width=0.9\linewidth]{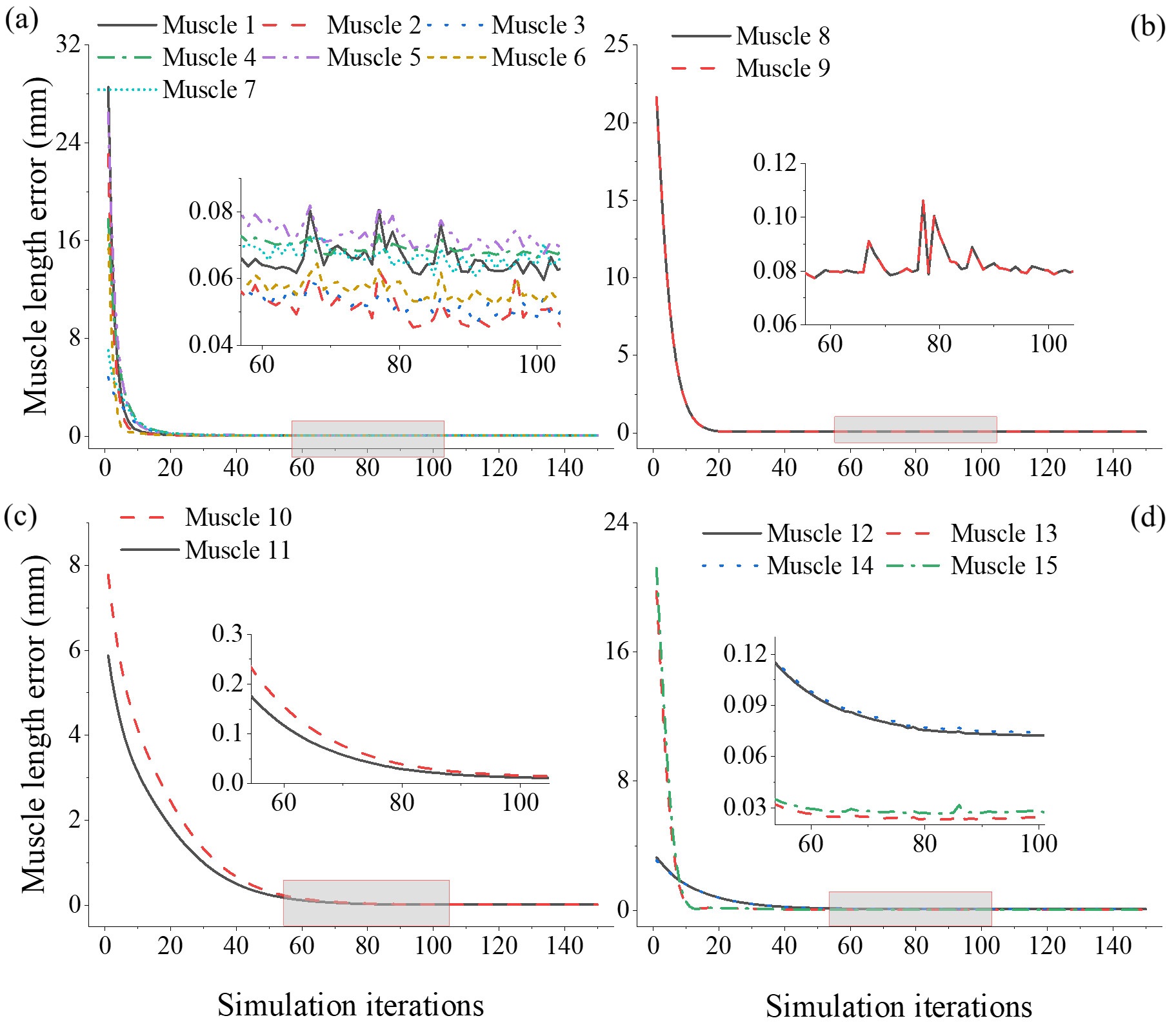}
    \caption{Simulation muscle length error results: (a) Shoulder joint data; (b) Elbow joint data; (c) Forearm joint data; (d) Wrist joint data}
\end{figure}

\begin{figure}[t]
	\centering
	\subfloat[Simulation trajectory]{
		\includegraphics[width=0.5\linewidth]{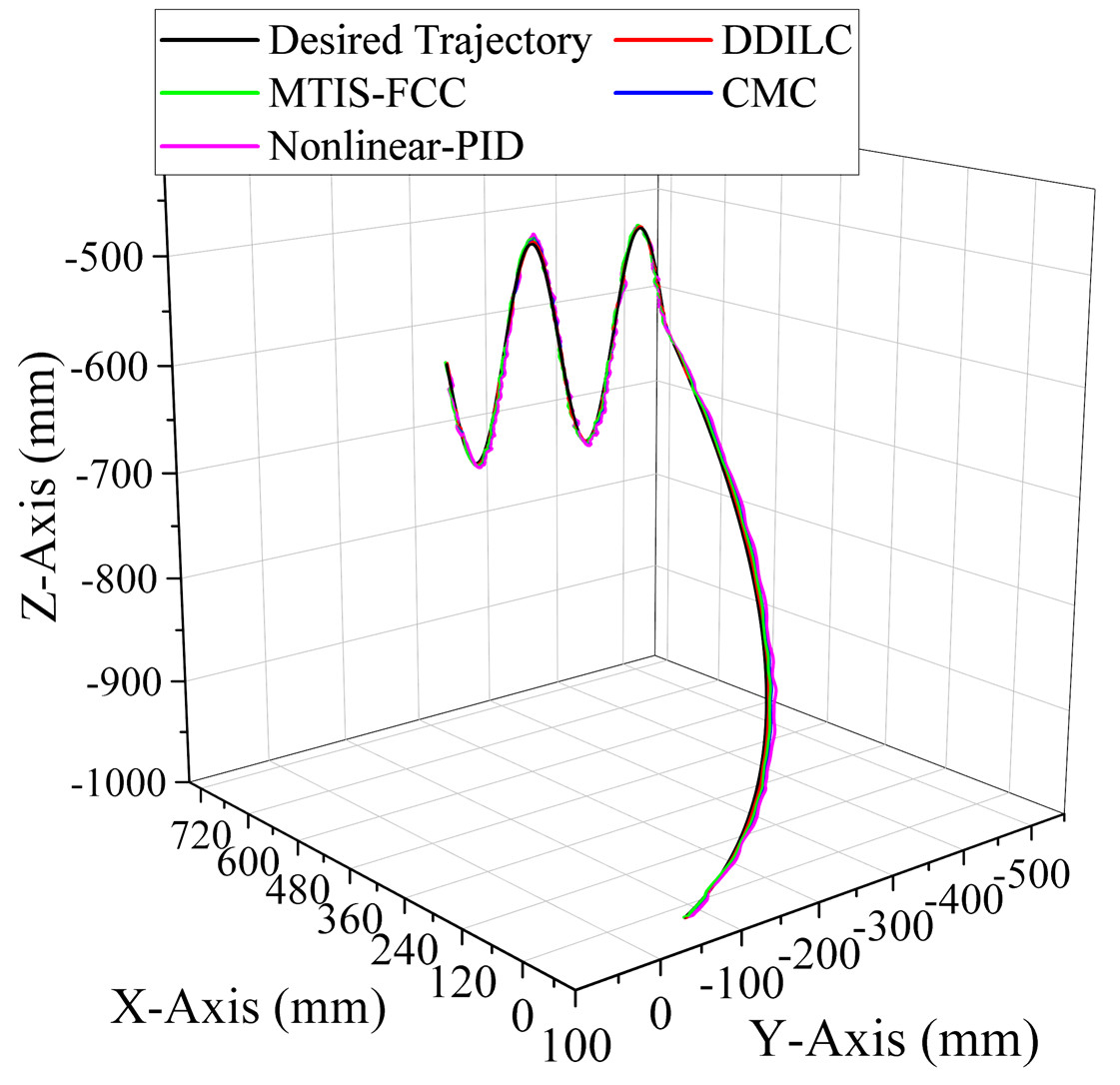}
	} \hfill
	\subfloat[Simulation trajectory error]{
		\includegraphics[width=0.45\linewidth]{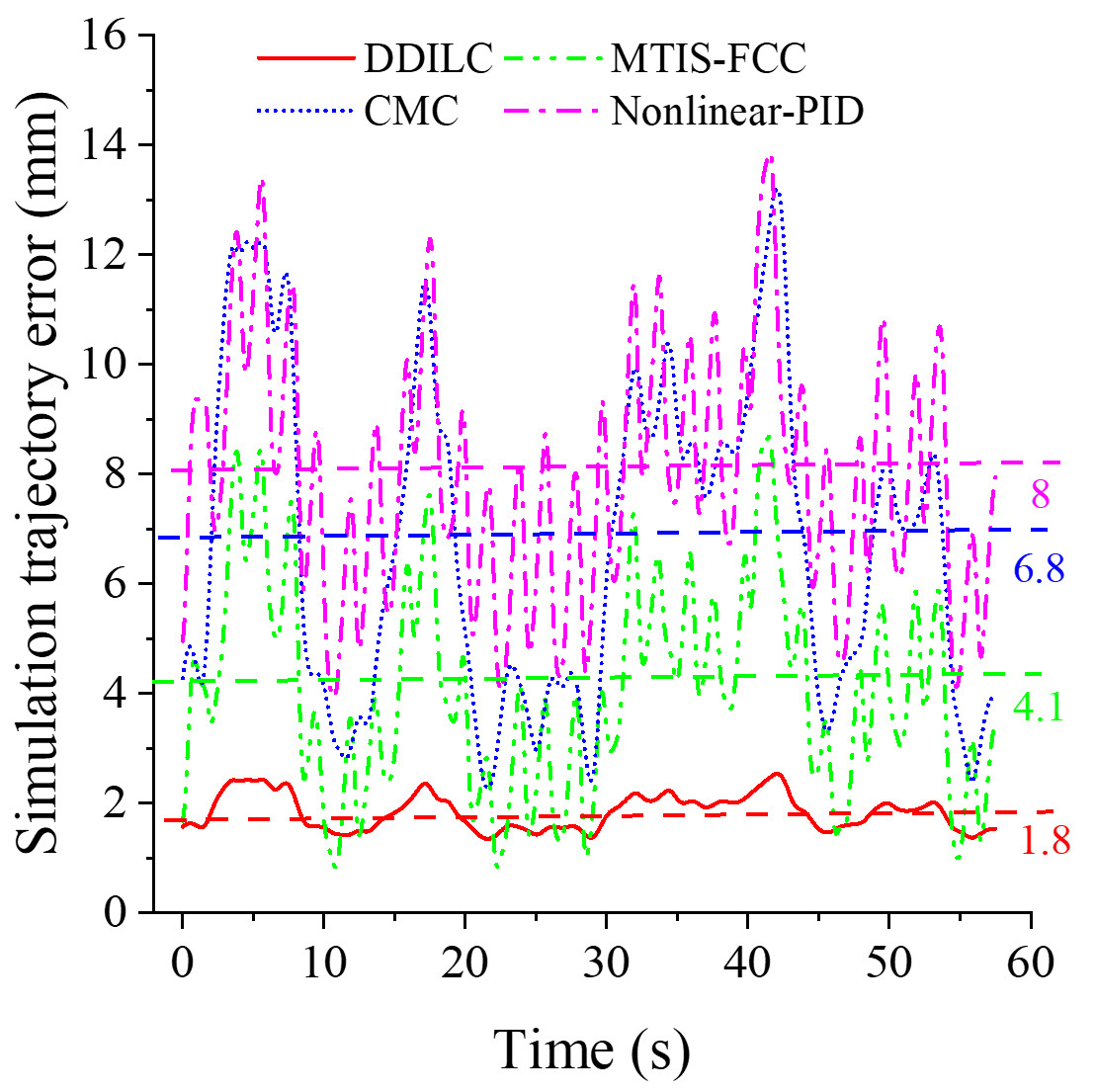}
	}
	\caption{Comparison of simulation results of DDILC and CMC of LTDM-Arm}
\end{figure}

\begin{figure}[t]
	\centering
	\subfloat[Trajectory under Different Disturbance Loads]{
		\includegraphics[width=0.5\linewidth]{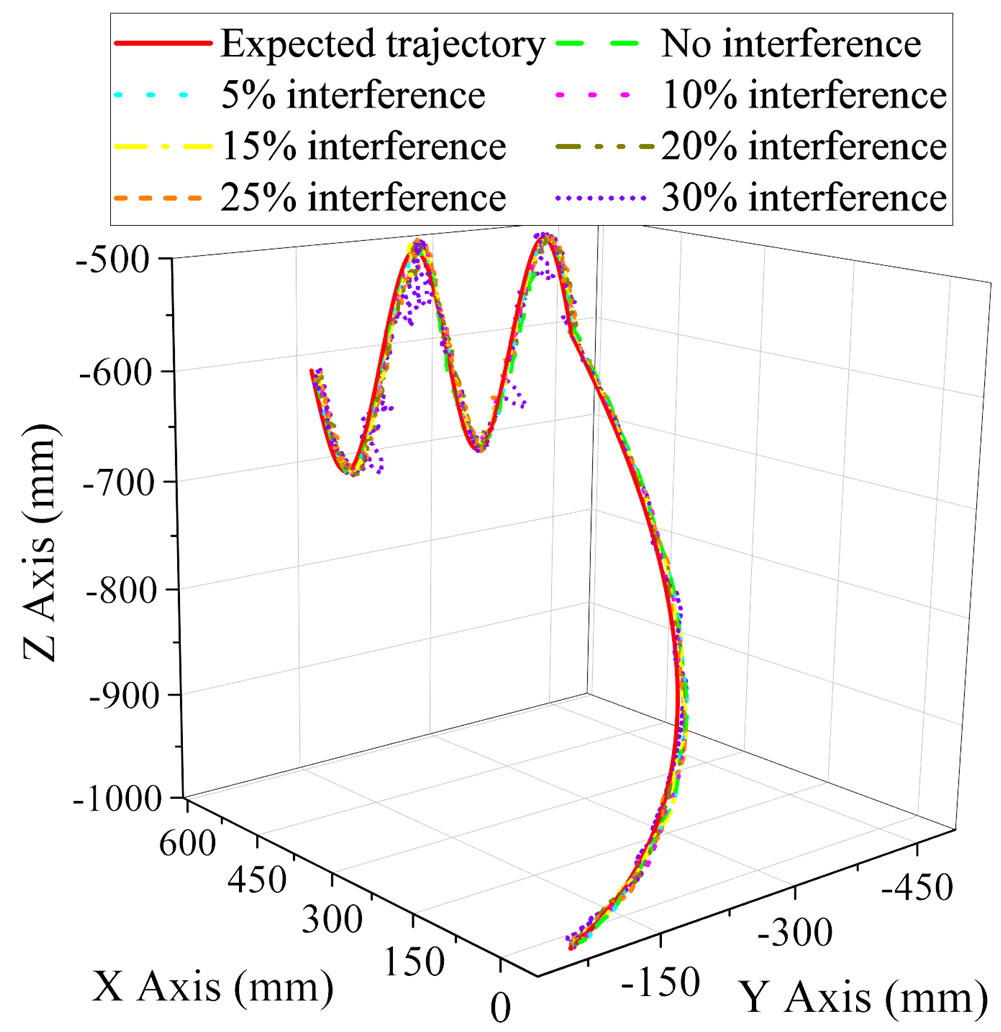}
	} \hfill
	\subfloat[Trajectory Error under Different Disturbance Loads]{
		\includegraphics[width=0.45\linewidth]{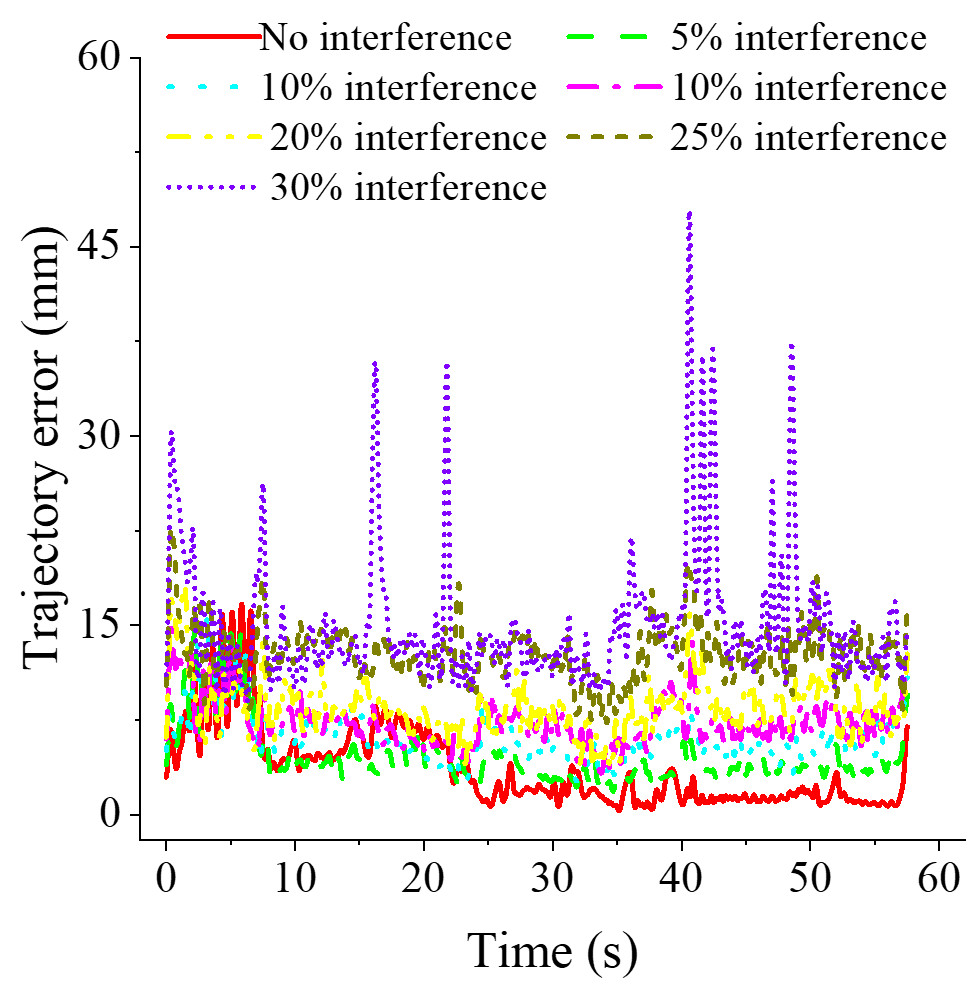}
	}
	\caption{Numerical Simulation Results for Interference Resistance}
\end{figure}

\begin{figure}[t]
	\centering
	\subfloat[Experimental trajectory]{
		\includegraphics[width=0.5\linewidth]{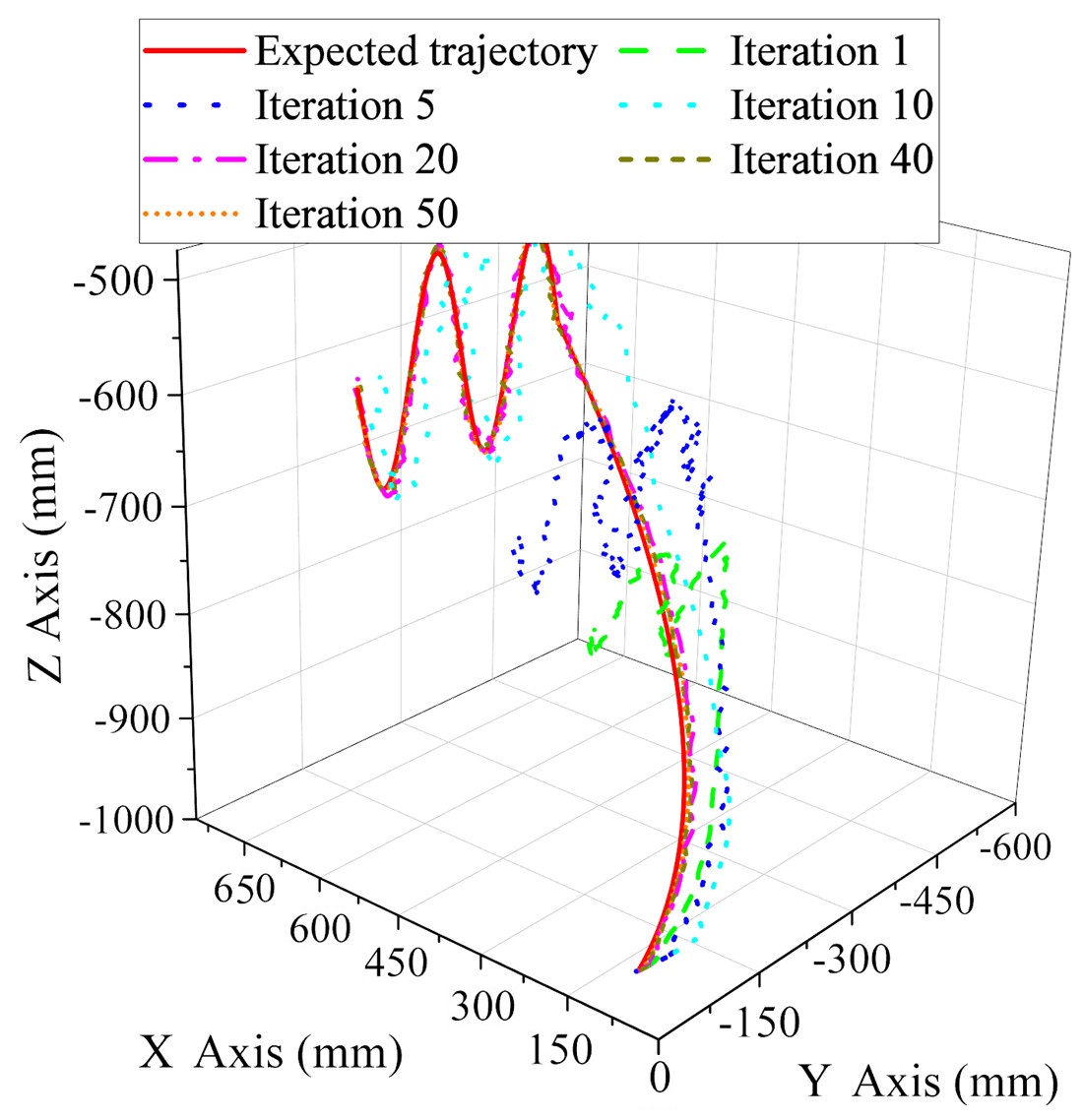}
	} \hfill
	\subfloat[Experimental trajectory error]{
		\includegraphics[width=0.45\linewidth]{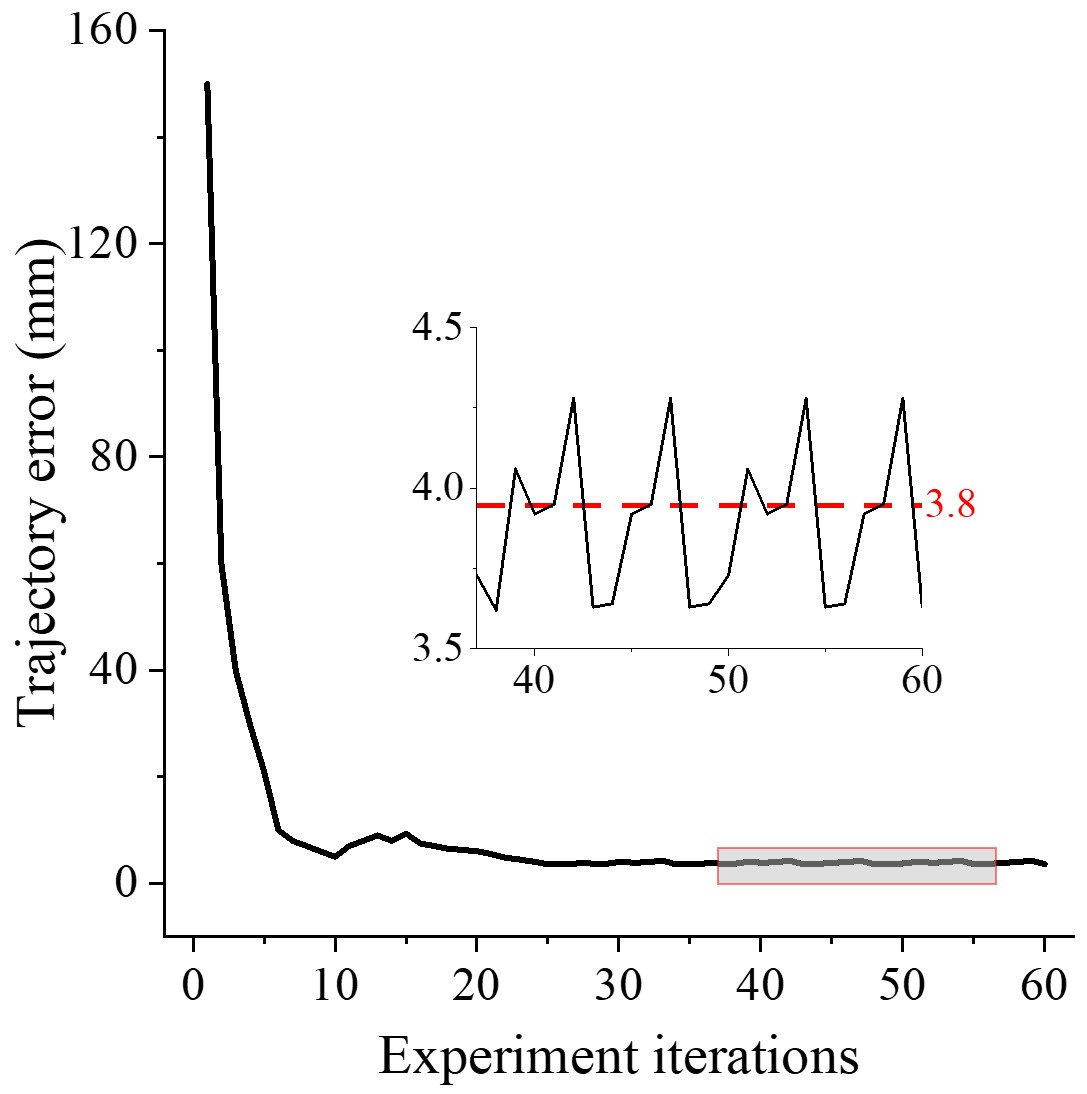}
	}
	\caption{Trajectory tracking results of the LTDM-Arm DDILC prototype experiment}
\end{figure}

\section*{5 Design and Results of Numerical Simulations and Experiments}

This section presents motion experiments based on the control methods outlined in Section 3, conducted in both the Mujoco-Python virtual simulation environment and with the LTDM-Arm prototype. A sine wave motion trajectory, spanning two cycles with a period and amplitude of 200 mm and 150 mm , respectively, was designed in a three-dimensional workspace. Each trial lasts 60 seconds and is repeated 10 times. Lastly, anti-disturbance experiments are performed in both the simulation and LTDM-Arm prototype to validate and analyze the results.

\subsection*{5.1 Numerical simulation results and analysis}

1) DDILC numerical simulation results and analysis

After 60 iterations, the average trajectory error decreases to below $2 \mathrm{~mm}(0.2 \%)$ and gradually stabilizes. Once the number of iterations exceeds 90 , the average error stabilizes around 1.38 mm ( $0.14 \%$ ), with a standard deviation of approximately 0.6 mm , as shown in Fig.6. The iteration errors of the muscle clusters for the four joints decrease progressively, with the average and mean square errors of the overall muscle lengths being 0.053 mm and 0.062 mm , respectively, as shown in Fig.7 . Overall, the numerical simulation trajectories closely align with the desired trajectories. However, due to the influence of multiple muscles acting on the same joint, initial errors among muscles are inconsistent. As iterations increase, antagonistic muscles help promote convergence, thereby accelerating the learning process. The direct control of muscle activation signals decouples the relationship between muscle space and joint space. The muscle model algorithm also features a small closed-loop adjustment mechanism, allowing the antagonistic relationship among muscles to stabilize rapidly and achieve high trajectory tracking accuracy.


2) Results and analysis of comparison with other control methods

Numerical simulations were conducted using the DDILC method, CMC method~\cite{8}, MTIS-FCC method~\cite{38} and Nonlinear-PID method. The average trajectory errors were approximately $1.8 mm (0.18 \%)$, $4.1 mm (0.42 \%)$, $6.8 mm (0.69 \%)$, and $8 mm(0.81 \%)$, respectively. The corresponding mean squared errors were $3.49 {mm}^2$, $20.46 {mm}^2$, $54.81 {mm}^2$, and $68.62 {mm}^2$, respectively. The proposed DDILC method demonstrates improvements of $56.1 \%, 73.53 \%$, and $77.5 \%$ compared to the baseline methods, as shown in Fig.8. Continuous learning and correction via the DDILC method lead to higher accuracy and effectively suppress error fluctuations. For the forearm joint, its unique structure causes significant changes in joint moment arms during motion, requiring precise muscle force adjustments. The DDILC method, through iterative learning and the dual effects of muscle model algorithms, compensates for the limitations of fixed gain adjustments and large proportional feedback control methods, thus enhancing the control accuracy of the LTDM-Arm under complex trajectory requirements.

3) Interference immunity of simulation results and analysis

Robot anti-interference manifests in two ways: stabilizing the system under disturbance and completing tasks effectively despite disturbances~\cite{18}. Due to model accuracy and workload errors, stable operation in unstructured environments is challenging. The LTDM-Arm demonstrates antiinterference through its unique driving mechanism and DDILC method. Trajectory tracking accuracy was analyzed under different disturbance loads, with amplitudes set at $0 \%$, $5 \%$, $10 \%$, $15 \%$, $20 \%$, $25 \%$, and $30 \%$ of the LTDM-Arm's rated 2.5 kg load. The simulation results (Fig.9) show average trajectory errors of $3.69 \mathrm{~mm}(0.38 \%)$, $4.53 \mathrm{~mm}(0.46 \%)$, $6.01 \mathrm{~mm}(0.61 \%)$, $7.31 \mathrm{~mm}(0.74 \%)$, $8.4 \mathrm{~mm}(0.86 \%)$, $12.8 \mathrm{~mm}(1.3 \%)$, and $15.05 \mathrm{~mm}(1.53 \%)$. Disturbance beyond $25 \%$ causes significant trajectory fluctuations, but errors remain under $1 \%$ when disturbance is less than $25 \%$, indicating strong load interference resistance. This anti-interference capability can be attributed to two factors:

(1) Special driving mechanism of the muscle model

The muscle activation signal affects the muscle fiber velocity, which in turn adjusts the tendon force. The muscle fiber length determines the tendon force, and this force, acting on the skeletal system, feeds back to adjust the muscle fiber length. This creates a small closed-loop control system where muscle fiber length and contraction speed serve as state feedback, and tendon force is the output. Internal feedback regulates muscle tension by adjusting muscle fiber length to correct errors and improve motion accuracy in the musculoskeletal robot. This process introduces a delay between input and output, known as muscle activation dynamics. The special coupling mechanism functions like a low-pass filter, effectively eliminating pulse or high-frequency interference in the activation signal.

(2) Coupling relationship between multiple antagonistic muscles:

The interactive force coupling among antagonistic muscles generates variable stiffness characteristics through co-contraction modulation. This inherent compliance enables the LTDM-Arm system to passively compensate for intrinsic oscillations and external disturbances, particularly when operating in unstructured environments with unknown loads and parametric uncertainties. The muscle synergy-driven stiffness adaptation mechanism demonstrates enhanced disturbance rejection capability, thereby improving environmental adaptability while maintaining dynamic stability.

\subsection*{5.2 LTDM-Arm prototype experimental results and analysis}

1) Analysis of DDILC experimental results

Following numerical simulations, trajectory tracking validation was implemented on the LTDMArm prototype using the DDILC strategy. As illustrated in Figs.10-11, the DDILC algorithm achieves progressive error reduction through iterative compensation of feedforward-feedback deviations. Post 40 iterations, the end-effector tracking error converges to $3.8 \mathrm{~mm}(0.4 \%)$. Compared with simulation benchmarks, the physical prototype exhibits heightened sensitivity to real-world uncertainties, including cumulative manufacturing tolerances, nonlinear joint friction, and sensor quantization effects. These empirical results confirm the DDILC's capability to mitigate system-level perturbations through adaptive learning.

\begin{figure}[t!]
	\centering
	\includegraphics[width=0.9\linewidth]{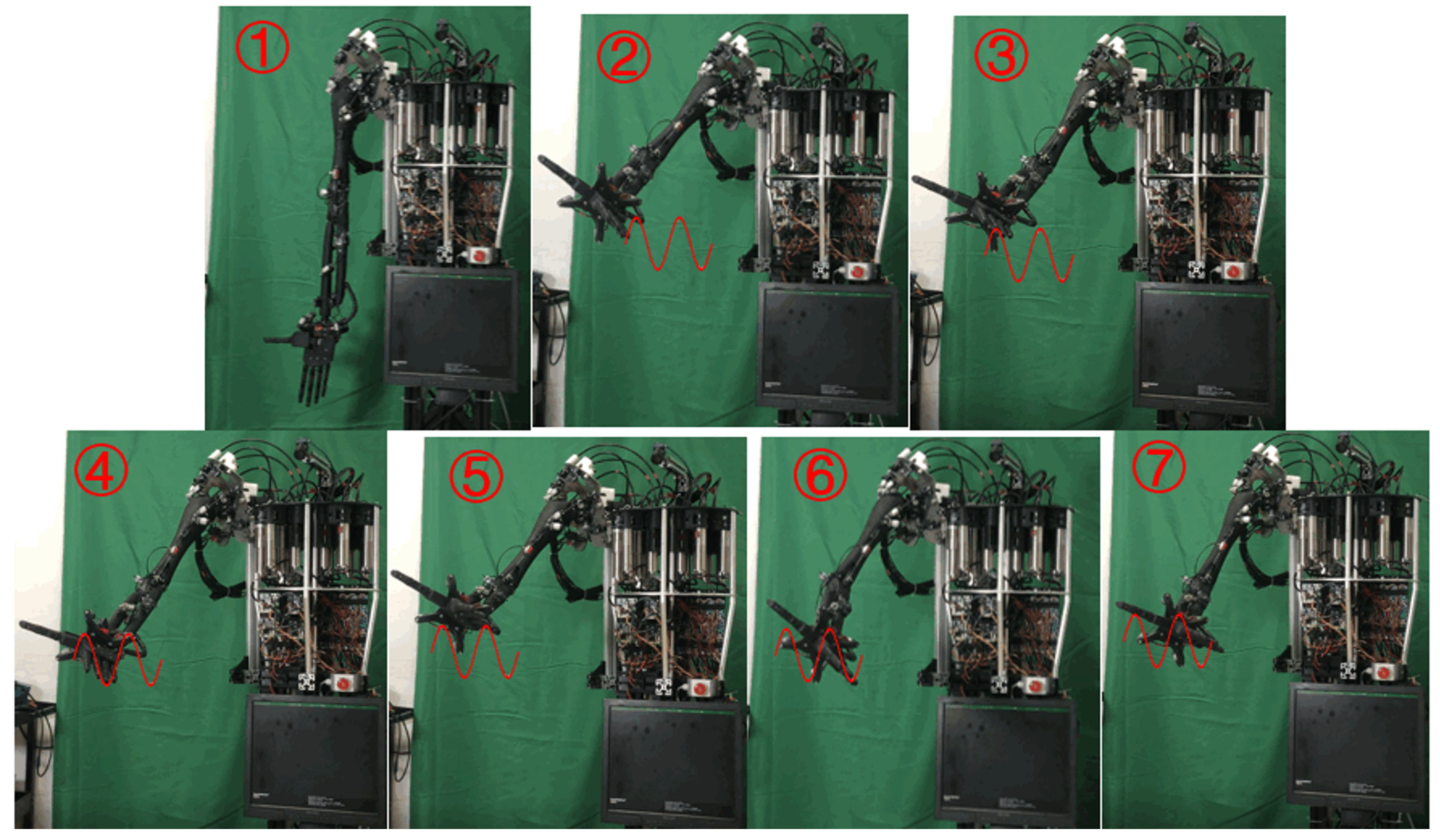}
	\caption{Experimental process of LTDM-Arm DDILC prototype}
\end{figure}

\begin{figure}[t!]
	\centering
	\subfloat[Experimental trajectory]{
		\includegraphics[width=0.5\linewidth]{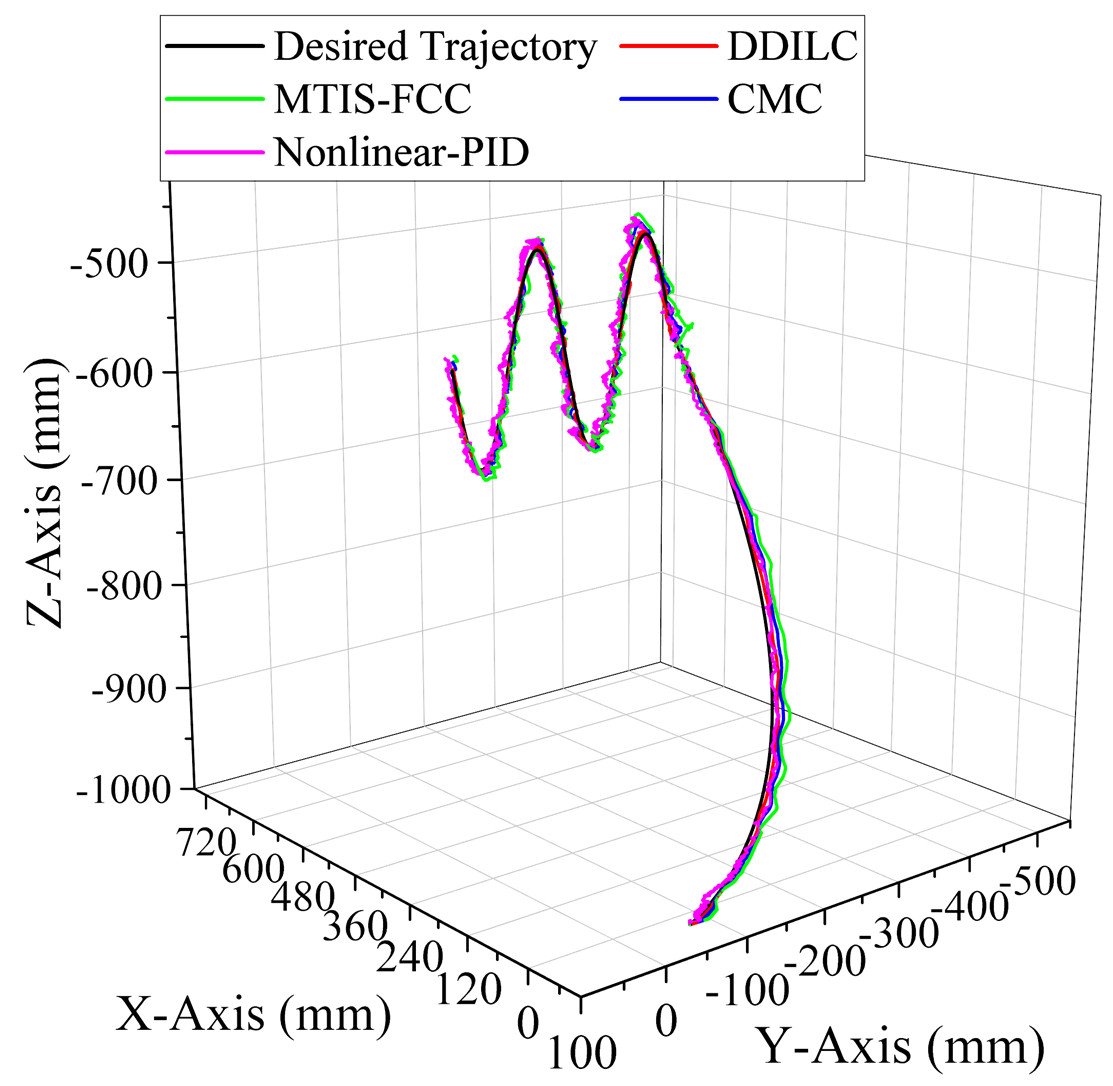}
	} \hfill
	\subfloat[Experimental trajectory error]{
		\includegraphics[width=0.45\linewidth]{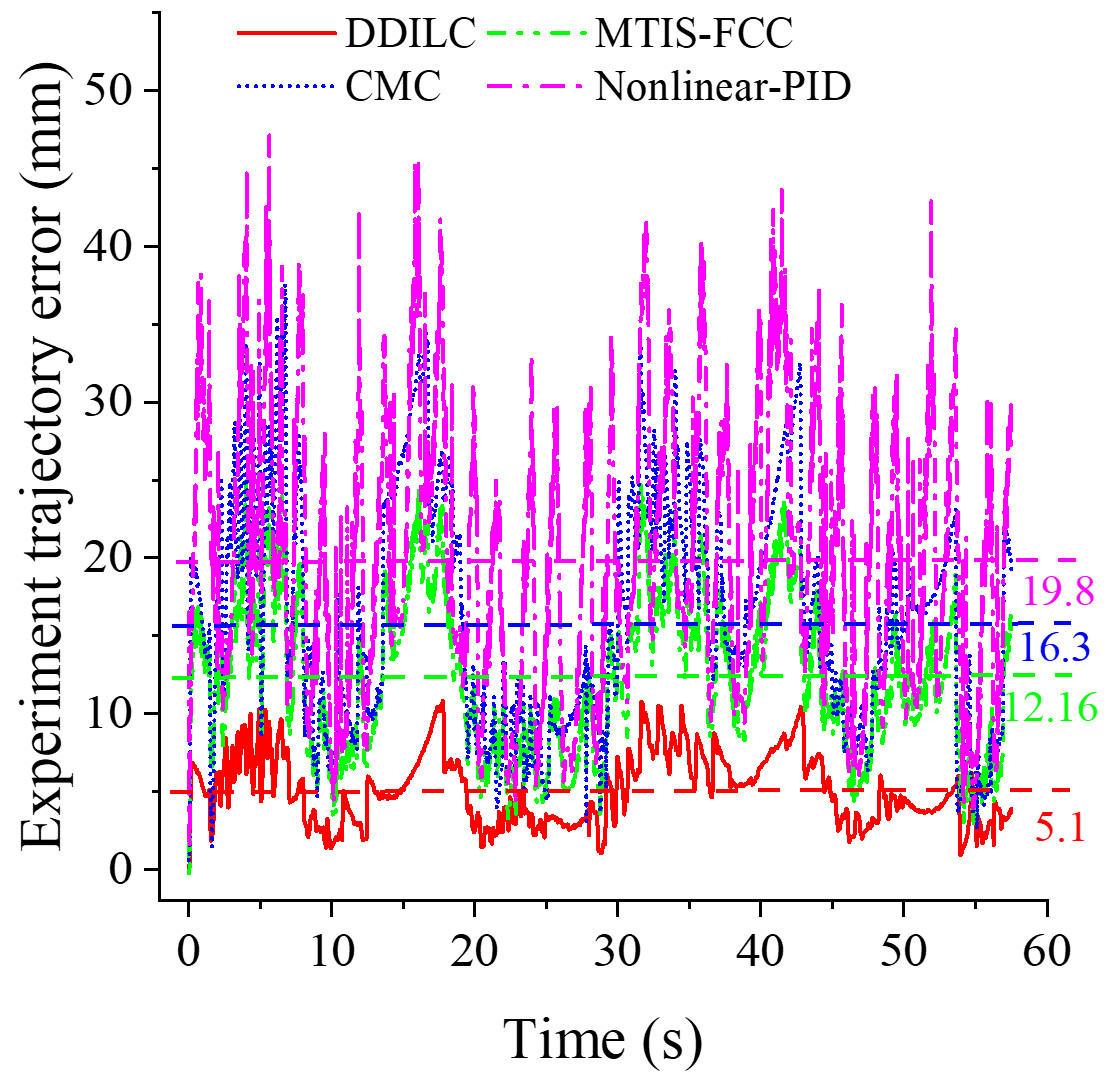}
	}
	\caption{Comparison of DDILC and CMC experimental results of LTDM-Arm prototype}
\end{figure}

\begin{figure}[t!]
	\centering
	\subfloat[Experimental Trajectory of the Prototype]{
		\includegraphics[width=0.5\linewidth]{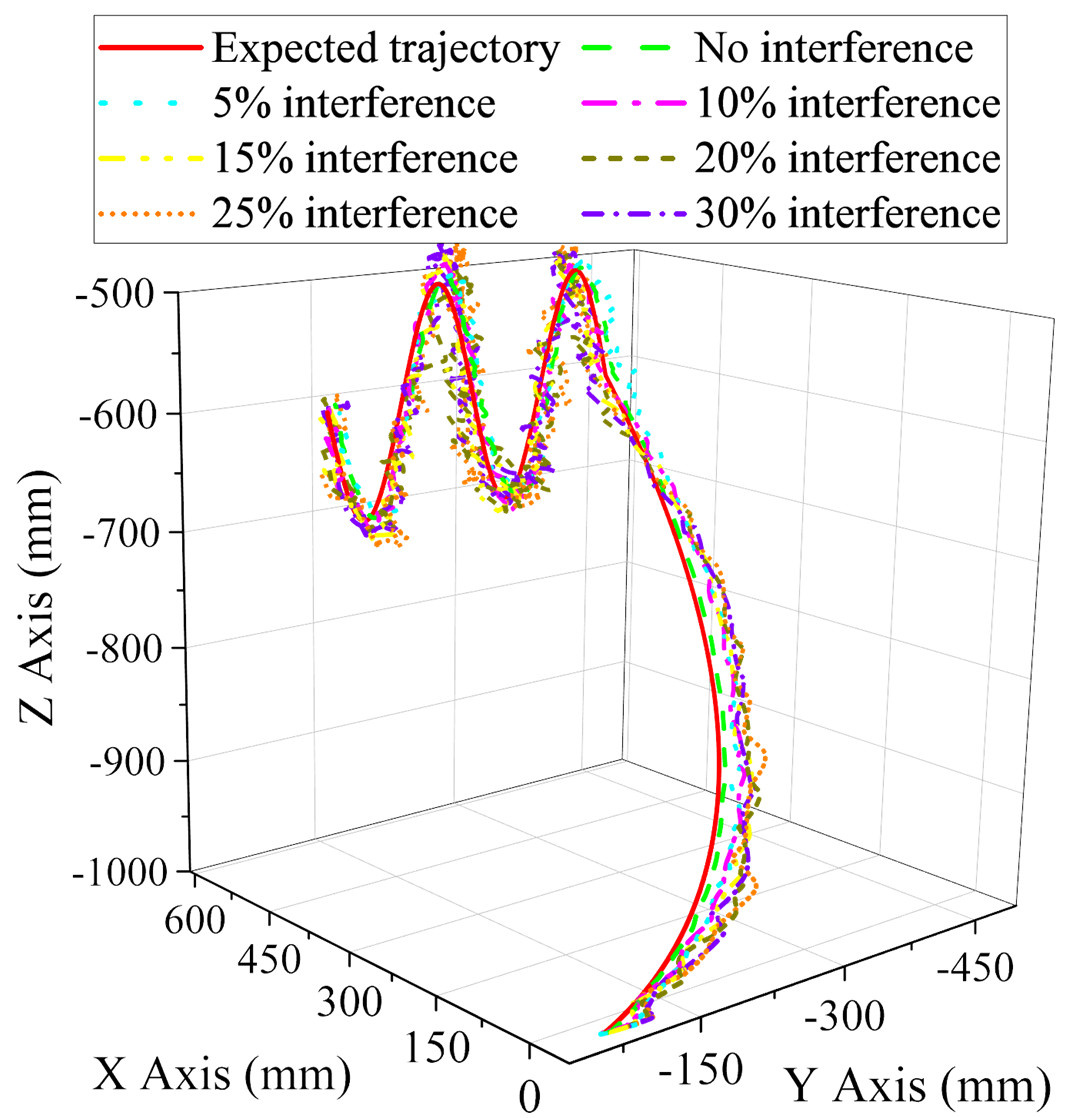}
	} \hfill
	\subfloat[Trajectory Error of the Prototype Experiment]{
		\includegraphics[width=0.45\linewidth]{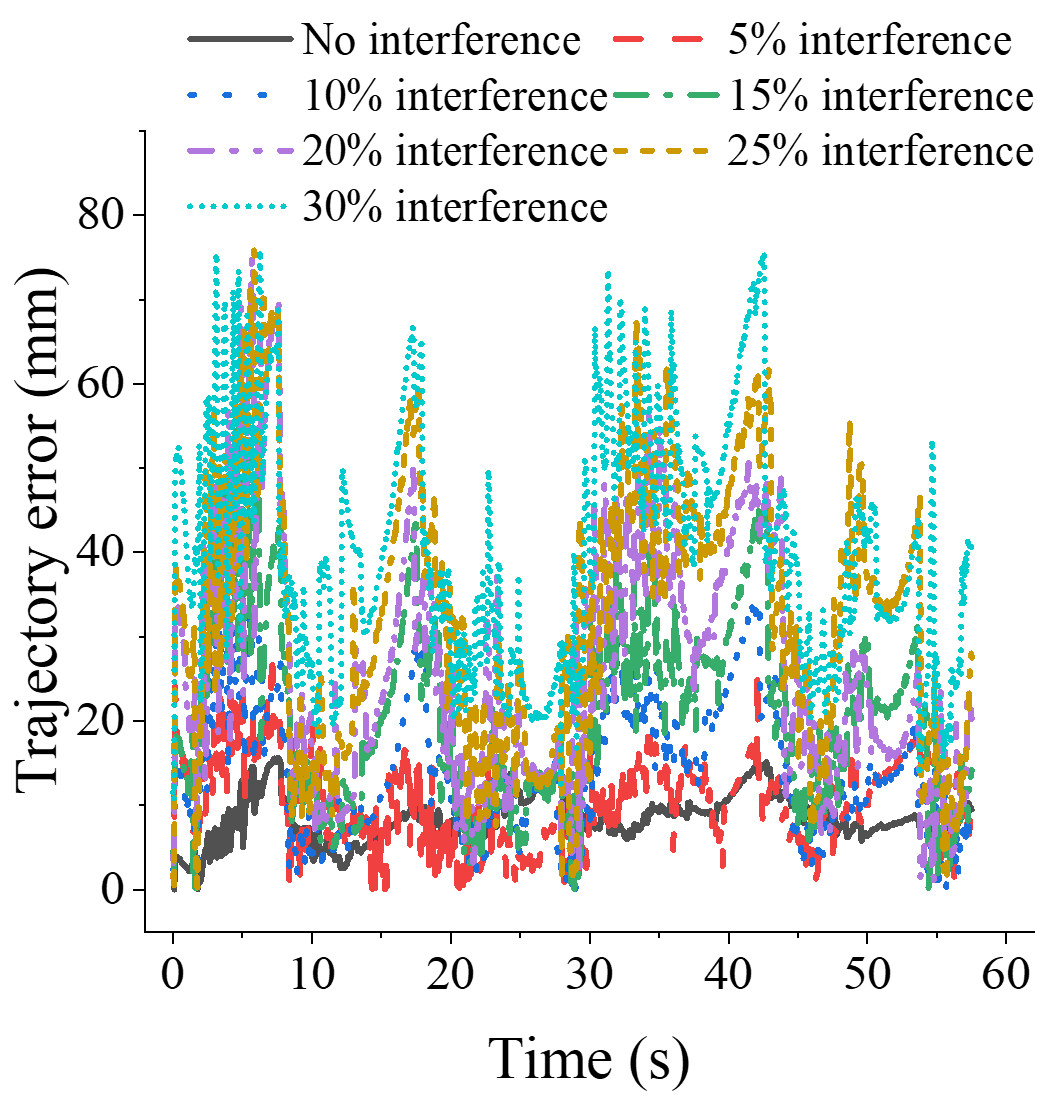}
	}
	\caption{Interference resistance experiment results of the LTDM-Arm prototype}
\end{figure}

2) Comparison and analysis of results with other control methods

Experimental evaluations of DDILC, computed muscle control (CMC), MTIS-FCC, and Nonlinear-PID methods yielded average trajectory tracking errors of $5.13 mm(0.52 \%)$, $12.16 mm ( 1.23 \%)$, $16.33 {mm}(1.66 \%)$, and $19.83 mm(2.13 \%)$, respectively, with corresponding mean squared errors (MSE) of $31.43 {mm}^2$, $173.2 {mm}^2$, $322.54 {mm}^2$, and $470.38 {mm}^2$, respectively. As demonstrated in Fig.12, the proposed DDILC method achieves significant error reduction rates of $57.8 \%, 68.59 \%$, and $74.13 \%$ relative to CMC, MTIS-FCC, and Nonlinear-PID baselines. These results empirically validate DDILC's superior precision in managing the LTDM-Arm's nonlinear dynamics and hardware-induced uncertainties. This can be attributed to several factors: First, the results align with numerical simulations. Second, the LTDM-Arm prototype is affected by manufacturing errors, sensor feedback inaccuracies, and cable transmission errors, making other methods more prone to overshooting. The DDILC method, however, refines the input activation signal through iterative adjustments and leverages the muscle model's closed-loop feedback, leading to more precise trajectory tracking. Experiments with the LTDM-Arm prototype confirm that the DDILC method is more robust to factors like large curvature changes and hardware friction.

3) Experiment results and analysis of interference resistance

During the manufacturing and installation of the LTDM-Arm prototype, model errors and friction are inevitable. However, the small closed-loop regulation mechanism of the muscle model, along with the antagonistic actions of multiple muscles, helps compensate for some of these hardware errors. In this section, the activation signal from the 40th iteration of the DDILC experiment on the LTDM-Arm prototype is used as the open-loop muscle activation input, with interference signals matching those used in the simulation experiments for the resistance test. The average trajectory errors observed were $8.43 \mathrm{~mm}(0.86 \%), 10.20 \mathrm{~mm}(1.04 \%), 14.22 \mathrm{~mm}(1.45 \%), 20.97 \mathrm{~mm}(2.14 \%), 26.44 \mathrm{~mm}(2.69 \%)$, $32.68 \mathrm{~mm} (3.33 \% )$, and $40.34 \mathrm{~mm} (4.11 \%)$, as shown in Fig.13. Due to model inaccuracies, the accuracy without load is lower than in the numerical simulation. When the interference load exceeds $15 \%$, the trajectory error surpasses $2 \%$, but the system still stabilizes. This further validates the LTDMArm's resistance to interference, as supported by the simulation analysis.

\section*{6 Conclusion}

Structures with human-like driving mechanisms are more likely to achieve human-like performance. To explore new musculoskeletal robotic systems, the LTDM-Arm was designed with seven DOF skeletal joints, retaining the humerus, ulna, and radius structures, and 15 MAMSs. To address challenges such as multi-redundant muscle antagonistic actuation, nonlinear muscle models, and complex rigid-flexible coupling systems, the DDILC and nonlinear muscle-modeling algorithms were applied to investigate the robustness of the LTDM-Arm. Experimental results of the LTDMArm's terminal-point trajectory tracking showed that: 1) The muscle model's internal regulation mechanism and multi-muscle antagonism make the LTDM-Arm robust to input signal variations; 2) The multi-muscle-driven single-joint structure generates a variable stiffness effect, allowing the LTDM-Arm to overcome joint friction and load inaccuracies. In open-loop input activation signal control simulations, the LTDM-Arm could still complete tasks with over $20 \%$ signal interference. Compared to the MTIS-FCC, CMC, and Nonlinear-PID methods on the LTDM-Arm platform, the DDILC method yielded smoother trajectories with smaller errors. These results demonstrate that the LTDM-Arm system is robust to input signal errors, hardware size inaccuracies, and friction. This study provides valuable experimental evidence for advancing anthropomorphic musculoskeletal robot development.

The LTDM-Arm has the potential to serve as a next-generation robotic system for repetitive labor tasks in unstructured environments. However, musculoskeletal robots still face challenges, particularly in safe and stable human-computer interaction. Humans rely on various feedback, such as vision, hearing, touch, and smell, to recognize their environment and generate muscle activation signals. Future work will focus on: 1) Exploring safe and stable human-computer interactions with the LTDMArm, and 2) Integrating vision, force sensing, and position sensing for autonomous multi-sensor fusion tasks, such as assembly work.

\Acknowledgements{This work is supported partly by Supported by the Strategic Priority Research Program of Chinese Academy of Science (Grant No. XDB32050100), the National Natural Science Foundation of China (under Grant 91948303).}



\section{Dynamic linearization convergence proof}

\textbf{Assumption 1:} Assume that the first-order partial derivative of the relative tracking error $\boldsymbol{e}(t+1, k), \ldots, \boldsymbol{e}\left(t-n_{e, t}+2, k\right)$ with respect to the ideal learning controller $\boldsymbol{C}(\bullet)$, where $0 \leq l_{e, t} \leq n_{e, t} \quad l_{e, t} \in \mathbb{Z}^{+}$.

\textbf{Assumption 2:} Assume that the ideal learning controller (17) satisfies the general Lipschitz condition , there exists a constant:
\begin{equation*}
\left\|\Delta \boldsymbol{u}^{b}(t, k)\right\| \leq \rho_{e}\|\Delta \boldsymbol{E}(t+1, k)\| \tag{27}
\end{equation*}
where,
$$
\begin{aligned}
& \boldsymbol{\Delta E}(t+1, k)=\left[\Delta \boldsymbol{e}(t+1, k), \boldsymbol{\Delta}(t, k), \ldots, \boldsymbol{\Delta}\left(t-l_{e, t}+2, k\right)\right]^{T} \in \mathbb{R}^{m\left(l_{e, t}\right)}, \\
& \boldsymbol{\Delta e}(t+1, k)=\boldsymbol{e}(t+1, k)-\boldsymbol{e}(t, k),
\end{aligned}
$$
$\rho_{e}$ is a positive constant, and Assumption 1 is the general condition of the learning controller, and Assumption 2 indicates that the ideal learning controller is an energy consumption unit and is stable.

\textbf{Theorem 1:} The ideal learning controller (17) satisfies Assumptions 1 and 2. If condition $\|\boldsymbol{E}(t+1, k)\| \neq 0$ is met. There exists a learning control gain parameter $\Xi_{e}(t, k) \in \mathbb{R}^{m\left(l_{e, t}\right)}$ that satisfies the pseudo partitioned Jacobian matrix (PPJM) condition so that the ideal feedback controller along the time axis is transformed into the following equivalent form:
\begin{equation*}
\Delta \boldsymbol{u}^{b}(t, k)=\boldsymbol{\Xi}_{e}^{T}(t, k) \boldsymbol{\Delta}(t+1, k) \tag{28}
\end{equation*}
where, $\boldsymbol{\Xi}_{e}(t, k)=\left[\boldsymbol{\Xi}_{1}(t, k), \ldots, \boldsymbol{\Xi}_{l_{e, t}}(t, k)\right] \in \mathbb{R}^{m \times m\left(l_{e, t}\right)}$ is the learning control gain vector parameter,

$$
\boldsymbol{\Xi}_{j}(t, k)=\left[\begin{array}{cccc}
\boldsymbol{\Xi}_{11 j}(t, k) & \boldsymbol{\Xi}_{12 j}(t, k) & \ldots & \boldsymbol{\Xi}_{1 m j}(t, k) \\
\boldsymbol{\Xi}_{2 l j}(t, k) & \boldsymbol{\Xi}_{12 j}(t, k) & \ldots & \boldsymbol{\Xi}_{2 m j}(t, k) \\
\vdots & \vdots & \ddots & \vdots \\
\boldsymbol{\Xi}_{m 1 j}(t, k) & \boldsymbol{\Xi}_{m 2 j}(t, k) & \cdots & \boldsymbol{\Xi}_{m m j}(t, k)
\end{array}\right] \in \mathbb{R}^{m \times m},
$$
$\left\|\boldsymbol{\Xi}_{e}(t, k)\right\| \leq b_{1}, \quad b_{1}>0$ is an unknown constant.

\textbf{Proof and analysis:}

The ideal learning controller $\boldsymbol{\Delta} \boldsymbol{u}^{b}(t, k)$ in (17) can be written as:
\begin{align*}
\Delta \boldsymbol{u}^{b}(t, k)= & \boldsymbol{C}\left(\boldsymbol{e}(t+1, k), \ldots, \boldsymbol{e}\left(t-n_{e, t}+2, k\right)\right) \\
& -\boldsymbol{C}\left(\boldsymbol{e}(t, k), \boldsymbol{e}(t, k), \ldots, \boldsymbol{e}\left(t-n_{e, t}+2, k\right)\right)+\boldsymbol{\Theta}_{1} \tag{29}
\end{align*}
where,
$$
\begin{aligned}
\boldsymbol{\Theta}_{1}= & \boldsymbol{C}\left(\boldsymbol{e}(t, k), \boldsymbol{e}(t, k), \ldots, \boldsymbol{e}\left(t-n_{e, t}+2, k\right)\right) \\
& -\boldsymbol{C}\left(\boldsymbol{e}(t, k), \ldots, \boldsymbol{e}\left(t-n_{e, t}+1, k\right)\right)
\end{aligned}
$$

According to Assumption 1 and the differential mean value theorem, formula (29) can be derived as follows:
\begin{equation*}
\Delta \boldsymbol{u}^{b}(t, k)=\frac{\partial \boldsymbol{C}^{*}}{\partial \boldsymbol{e}(t+1, k)} \boldsymbol{\Delta} \boldsymbol{e}(t+1, k)+\boldsymbol{\Theta}_{1} \tag{30}
\end{equation*}

where,
$$
\frac{\partial \boldsymbol{C}^{*}}{\partial \boldsymbol{e}(t+1, k)}=\left[\begin{array}{cccc}
\frac{\partial \boldsymbol{C}_{1}^{*}}{\partial e_{1}(t+1, k)} & \frac{\partial \boldsymbol{C}_{1}^{*}}{\partial e_{2}(t+1, k)} & \cdots & \frac{\partial \boldsymbol{C}_{1}^{*}}{\partial e_{m}(t+1, k)} \\
\frac{\partial \boldsymbol{C}_{2}^{*}}{\partial e_{1}(t+1, k)} & \frac{\partial \boldsymbol{C}_{2}^{*}}{\partial e_{2}(t+1, k)} & \cdots & \frac{\partial \boldsymbol{C}_{2}^{*}}{\partial e_{m}(t+1, k)} \\
\vdots & \vdots & \ddots & \cdots \\
\frac{\partial \boldsymbol{C}_{m}^{*}}{\partial e_{1}(t+1, k)} & \frac{\partial \boldsymbol{C}_{m}^{*}}{\partial e_{2}(t+1, k)} & \cdots & \frac{\partial \boldsymbol{C}_{m}^{*}}{\partial e_{m}(t+1, k)}
\end{array}\right]^{T} \in \mathbb{R}^{m \times m}
$$

$\partial C_{p}^{*} / \partial e_{q}(t+1, k)$ represents the value $\boldsymbol{C}(\bullet)$ of the first-order partial derivative of function $e_{q}(t+1, k)$ with respect to variable $\left[e_{q}(t+1, k), e_{q}(t, k)\right]$ at a point in the interval, $p=1,2, \ldots, m ; q=1,2, \ldots, m$.

Similarly, according to Assumption 1 and the mean value theorem, formula (30) is transformed into:

\begin{align*}
\Delta \boldsymbol{u}^{b}(t, k)= & \frac{\partial \boldsymbol{C}^{*}}{\partial \boldsymbol{e}(t+1, k)} \boldsymbol{\Delta}\boldsymbol{e}(t+1, k)+\frac{\partial \boldsymbol{\Theta}_{1}^{*}}{\partial \boldsymbol{e}(t, k)} \boldsymbol{\Delta} \boldsymbol{e}(t, k) \\
& +\boldsymbol{C}\left(\boldsymbol{e}(t, k), \boldsymbol{e}(t-1, k), \boldsymbol{e}(t-1, k), \ldots, \boldsymbol{e}\left(t-n_{e, t}+2, k\right)\right)  \tag{31}\\
& -\boldsymbol{C}\left(\boldsymbol{e}(t, k), \boldsymbol{e}(t-1, k), \boldsymbol{e}(t-2, k), \boldsymbol{e}(t-2, k), \ldots, \boldsymbol{e}\left(t-n_{e, t}+2, k\right)\right)+\boldsymbol{\Theta}_{4}
\end{align*}
where,
$$
\frac{\partial \boldsymbol{\Theta}_{1}^{*}}{\partial \boldsymbol{e}(t, k)}=\left[\begin{array}{cccc}
\frac{\partial \boldsymbol{\Theta}_{11}^{*}}{\partial e_{1}(t, k)} & \frac{\partial \boldsymbol{\Theta}_{11}^{*}}{\partial e_{2}(t, k)} & \cdots & \frac{\partial \boldsymbol{\Theta}_{11}^{*}}{\partial e_{m}(t, k)} \\
\frac{\partial \boldsymbol{\Theta}_{12}^{*}}{\partial e_{1}(t, k)} & \frac{\partial \boldsymbol{\Theta}_{12}^{*}}{\partial e_{2}(t, k)} & \cdots & \frac{\partial \boldsymbol{\Theta}_{12}^{*}}{\partial e_{m}(t, k)} \\
\vdots & \vdots & \ddots & \vdots \\
\frac{\partial \boldsymbol{\Theta}_{1 m}^{*}}{\partial e_{1}(t, k)} & \frac{\partial \boldsymbol{\Theta}_{1 m}^{*}}{\partial e_{2}(t, k)} & \cdots & \frac{\partial \boldsymbol{\Theta}_{1 m}^{*}}{\partial e_{m}(t, k)}
\end{array}\right]^{T} \in \mathbb{R}^{m \times m}
$$

$\partial \boldsymbol{\Theta}_{1 p}^{*} / \partial e_{q}(t, k)$ represents the value $\boldsymbol{C}(\bullet)$ of the first-order partial derivative of function $e_{q}(t, k)$\\
with respect to variable $\left[e_{q}(t+1, k), e_{q}(t, k)\right]$ at a point in the interval, $p=1,2, \ldots, m ; q=1,2, \ldots, m$.\\

Summarizing (29) to (31), we get:
\begin{equation*}
\Delta \boldsymbol{u}^{b}(t, k)=\left[\frac{\partial \boldsymbol{C}^{*}}{\partial \boldsymbol{e}(t+1, k)}, \frac{\partial \boldsymbol{\Theta}_{1}}{\partial \boldsymbol{e}(t, k)}, \ldots, \frac{\partial \boldsymbol{\Theta}_{l_{e, t}-1}}{\partial \boldsymbol{e}\left(t-l_{e, t}+2, k\right)}\right] \boldsymbol{\Delta}\boldsymbol{E}(t+1, k)+\boldsymbol{\Theta}_{l_{t a}} \tag{32}
\end{equation*}

Considering the following vector variable $\boldsymbol{\beta}_{\boldsymbol{e}}(t, k)$ for formula (32), we can obtain the numerical equation:

\begin{equation*}
\boldsymbol{\Theta}_{\boldsymbol{e}_{e, t}}=\boldsymbol{\beta}_{\boldsymbol{e}}^{T}(t, k) \Delta \boldsymbol{E}(t+1, k) \tag{33}
\end{equation*}

Combined with formula (28), due to $\Delta \boldsymbol{E}(t+1, k) \neq 0$, formula (33) has at least one solution $\boldsymbol{\beta}_{e}^{*}(t, k)$ such that:

\begin{equation*}
\boldsymbol{\Xi}_{e}(t, k)=\left[\frac{\partial \boldsymbol{C}^{*}}{\partial \boldsymbol{e}(t+1, k)}, \frac{\partial \boldsymbol{\Theta}_{1}}{\partial \boldsymbol{e}(t, k)}, \ldots, \frac{\partial \boldsymbol{\Theta}_{\boldsymbol{l}_{e t}-1}}{\partial \boldsymbol{e}\left(t-l_{e, t}+2, k\right)}\right]+\boldsymbol{\beta}_{e}^{*}(t, k) \tag{34}
\end{equation*}

According to Assumption 1, there exists a bounded constant such that $\left\|\boldsymbol{\Xi}_{e}(t, k)\right\| \leq b_{1}$, and the proof is complete.

\section{Controller convergence proof}

\textbf{Assumption 3:} Assume the following conditions exist:
\begin{equation*}
\|\boldsymbol{u}(t, k)\|_{\lambda}=\sup _{t[0, T]} \alpha^{-\lambda t}\|\boldsymbol{u}(t, k)\| \tag{35}
\end{equation*}
where, $\lambda>0, \alpha>1$, The following convergence is proved under this condition.

\textbf{Theorem 2:} Under assumptions 1, 2, and 3, for system (16) under the control methods (18), (22) to (25), if $\left\|1-\boldsymbol{\beta} \boldsymbol{f}_{u}(\boldsymbol{\varphi}(t, k))\right\|<1$ is satisfied, then $\lim _{k \rightarrow \infty}\left\|\delta \boldsymbol{u}^{f}(t, k)\right\|_{\lambda} \leq \boldsymbol{\sigma}, \lim _{k \rightarrow \infty}\|\boldsymbol{e}(t, k)\|_{\lambda} \leq \boldsymbol{\sigma}$, where, $\sigma=\frac{\varepsilon_{1} \varepsilon_{4}}{1-\boldsymbol{\rho}}>0$ is a constant. In addition, when the conditions $\|\boldsymbol{e}(0, k)\|=0,\left\|\boldsymbol{u}^{b}(0, k)\right\|=0$ and $\boldsymbol{y}_{\boldsymbol{d}}(t)$ are constants, there are $\lim _{k \rightarrow \infty}\left\|\delta \boldsymbol{u}^{f}(t, k)\right\|_{\lambda}=0$ and $\lim _{k \rightarrow \infty}\|\boldsymbol{e}(t, k)\|_{\lambda}=0$.

\textbf{Proof and analysis:}

The convergence of feedback dynamic linearization control is an open problem. This paper provides one of the convergence proof methods.
\begin{align*}
& \boldsymbol{e}(t+1, k) \\
& =\boldsymbol{y}_{\boldsymbol{d}}(t+1, k)-\boldsymbol{y}(t+1, k) \\
& =f\left(\boldsymbol{y}_{\boldsymbol{d}}(t), \boldsymbol{u}_{\boldsymbol{d}}(t)\right)-f\left(\boldsymbol{y}_{d}(t), \boldsymbol{u}(t, k)\right)+f\left(\boldsymbol{y}_{\boldsymbol{d}}(t), \boldsymbol{u}(t, k)\right)-f(\boldsymbol{y}(t, k), \boldsymbol{u}(t, k))  \tag{36}\\
& =f_{y}(\boldsymbol{\varphi}(t, k)) \boldsymbol{e}(t, k)+f_{u}(\varphi(t, k)) \boldsymbol{\delta} \boldsymbol{u}(t, k) \\
& =f_{u}(\boldsymbol{\varphi}(t, k)) \boldsymbol{\delta} \boldsymbol{u}^{f}(t, k)-f_{u}(\varphi(t, k)) \boldsymbol{u}^{b}(t, k)+f_{y}(\varphi(t, k)) \boldsymbol{e}(t, k)
\end{align*}
where, $\boldsymbol{\delta} \boldsymbol{u}(t, k)=\boldsymbol{u}_{\boldsymbol{d}}(t)-\boldsymbol{u}(t, k), \quad \boldsymbol{\varphi}(t, k)=\left[(\boldsymbol{u}(t, k)+\boldsymbol{\imath} \boldsymbol{\delta} \boldsymbol{u}(t, k))^{T},(\boldsymbol{e}(t, k)+\boldsymbol{\imath}(t, k))^{T}\right]^{T}, \boldsymbol{l} \in[0,1]$.\\

Combining formula (36), taking the norm on both sides of formula (25) yields:
\begin{align*}
& \left\|\boldsymbol{\delta} \mathbf{u}^{f}(t, k)\right\| \\
& =\left\|\boldsymbol{u}_{d}(t)-\boldsymbol{u}^{f}(t, k)\right\| \\
& =\left\|\boldsymbol{u}_{d}(t)-\boldsymbol{u}^{f}(t, k-1)-\boldsymbol{\beta} \boldsymbol{e}(t+1, k-1)\right\|  \tag{37}\\
& =\left\|\boldsymbol{\delta} \mathbf{u}^{f}(t, k-1)-\boldsymbol{\beta}\left(F_{u} \boldsymbol{\delta} \mathbf{u}^{f}(t, k-1)-F_{u} \boldsymbol{u}^{b}(t, k-1)+F_{y} \boldsymbol{e}(t, k-1)\right)\right\| \\
& =\left\|I-\boldsymbol{\beta} F_{u}\right\|\left\|\boldsymbol{\delta} \boldsymbol{u}^{f}(t, k-1)\right\|+\boldsymbol{\varepsilon}_{1}\left[\left\|\boldsymbol{u}^{b}(t, k-1)\right\|+\|\boldsymbol{e}(t, k-1)\|\right]
\end{align*}
where, $F_{u}=\sup _{t \in[0, T]}\left\|f_{u}(\varphi(t, k-1))\right\|, \quad F_{y}=\sup _{t \in[0, T]}\left\|f_{y}(\varphi(t, k-1))\right\|, \quad \varepsilon_{1}=\max _{t \in[0, T]}\left(\beta F_{u}, \boldsymbol{\beta} F_{y}\right)$.\\

Similarly, taking the norm on both sides of formula (36) yields:
\begin{equation*}
\|\boldsymbol{e}(t+1, k-1)\| \leq F_{u}\left\|\boldsymbol{\delta} \boldsymbol{u}^{f}(t-1, k-1)\right\|+F_{u}\left\|\boldsymbol{u}^{b}(t-1, k-1)\right\|+F_{y}\|\boldsymbol{e}(t-1, k-1)\| \tag{38}
\end{equation*}

For the feedback term $\boldsymbol{u}^{\boldsymbol{b}}(t, k)$, combining the feedback control rate (28) and (38), we get:
\begin{align*}
\left\|\boldsymbol{u}^{b}(t, k-1)\right\| 
&\leq \left\|\boldsymbol{u}^{b}(t-1, k-1)\right\| + \left\|\hat{\boldsymbol{\Xi}}_{e}^{T}(t, k-1)\right\| \|\boldsymbol{\Delta E}(t+1, k-1)\| \\
&\leq \left\|\boldsymbol{u}^{b}(t-1, k-1)\right\| + \boldsymbol{\varepsilon}_{2} \left\|\left[\Delta e(t+1, k-1), \ldots, \Delta e\left(t-l_{e, t}+2, k-1\right)\right]\right\| \\
&\leq \left\|\boldsymbol{u}^{b}(t-1, k-1)\right\| + \boldsymbol{\varepsilon}_{2} \left\|\boldsymbol{e}(t+1, k-1) - e\left(t-l_{e, t}+1, k-1\right)\right\| \tag{39} \\
&\leq \left\|\boldsymbol{u}^{b}(t-1, k-1)\right\| + 
\boldsymbol{\varepsilon}_2 \left\| 
\begin{array}{c}
F_u\left(\left\|\boldsymbol{\delta u}^f(t,k-1)\right\|-\left\|\boldsymbol{\delta u}^f(t-l_{e,t},k-1)\right\|\right)\\
-F_u\left(\left\|\boldsymbol{u}^b(t,k-1)\right\|-\left\|\boldsymbol{u}^b(t-l_{e,t},k-1)\right\|\right)\\
+F_y\left(\left\|\boldsymbol{e}(t,k-1)\right\|-\left\|\boldsymbol{e}(t-l_{e,t},k-1)\right\|\right)
\end{array}
\right\|
\end{align*}

where, $\boldsymbol{\varepsilon}_{2}=\sup _{t \in[0, T]}\left\|\boldsymbol{\Xi}_{e}^{T}(t, k-1)\right\|, \quad \boldsymbol{\Xi}_{e}^{T} \leq b_{1}, \quad b_{1}>0$ is an unknown constant.

Adding formula (38) and formula (39) yields:
\begin{align*}
& \left\|\boldsymbol{u}^{b}(t, k-1)\right\|+\|\boldsymbol{e}(t, k-1)\| \\
& \leq \boldsymbol{\varepsilon}_{3}\left(\begin{array}{l}
\left\|\boldsymbol{u}^{b}(t, k-1)\right\|+\left\|\boldsymbol{u}^{b}(t-1, k-1)\right\|+\left\|\boldsymbol{u}^{b}\left(t-l_{e, t}, k-1\right)\right\|+\|\boldsymbol{e}(t, k-1)\| \\
+\quad(t-k-1)\|+\| \boldsymbol{e}\left(t-l_{e, t}, k-1\right)\|+\| \boldsymbol{\delta} \boldsymbol{u}^{f}(t, k-1) \| \\
+\left\|\boldsymbol{\delta} \boldsymbol{u}^{f}(t-1, k-1)\right\|+\left\|\boldsymbol{\delta} \boldsymbol{u}^{f}\left(t-l_{e, t}, k-1\right)\right\|
\end{array}\right)  \tag{40}\\
& \leq\left(\boldsymbol{\varepsilon}_{3}^{t}+\boldsymbol{\varepsilon}_{3}^{t-1}+\boldsymbol{\varepsilon}_{3}^{t-l_{e, t}}\right)\left(\left\|\boldsymbol{u}^{b}(0, k-1)\right\|+\|\boldsymbol{e}(0, k-1)\|\right) \\
& +\sum_{j=0}^{T-1}\left(\left(\boldsymbol{\varepsilon}_{3}^{T-j}+\boldsymbol{\varepsilon}_{3}^{T-j-1}+\boldsymbol{\varepsilon}_{3}^{T-j-l_{e, t}, t}\right)\left\|\boldsymbol{\delta} \boldsymbol{u}^{f}(j, k-1)\right\|\right)
\end{align*}
where, $\varepsilon_{3}=\max _{t[0, T]}\left(1+F_{u}, F_{u}, F_{y}, \varepsilon_{2} F_{u},-\varepsilon_{2} F_{u}, \varepsilon_{2} F_{u}, \varepsilon_{2} F_{y},-\varepsilon_{2} F\right), \quad \varepsilon_{3}>1$.\\
Substituting formula (40) into formula (37) yields:
\begin{align*}
& \left\|\boldsymbol{\delta} \boldsymbol{u}^{f}(t, k)\right\| \\
& \leq\left(I-\boldsymbol{\beta} F_{u}\right)\left\|\boldsymbol{\delta} \mathbf{u}^{f}(t, k-1)\right\|+\boldsymbol{\varepsilon}_{1} \boldsymbol{\varepsilon}_{4}  \tag{41}\\
& +\sum_{j=0}^{T-1}\left(\left(\boldsymbol{\varepsilon}_{3}^{T-j}+\boldsymbol{\varepsilon}_{3}^{T-j-1}+\boldsymbol{\varepsilon}_{3}^{T-j-l_{e, t}+1}\right)\left\|\boldsymbol{\delta} \mathbf{u}^{f}(j, k-1)\right\|\right)
\end{align*}
where, $\boldsymbol{\varepsilon}_{4}=\left(\varepsilon_{3}^{t}+\varepsilon_{3}^{t-1}+\boldsymbol{\varepsilon}_{3}^{t-l_{e, t}}\right)\left(\left\|\boldsymbol{u}^{b}(0, k-1)\right\|+\|\boldsymbol{e}(0, k-1)\|\right)$.\\

According to Assumption 3, taking the $\lambda$ norm on both sides of formula (41) yields:
\begin{align*}
& \left\|\boldsymbol{\delta} \boldsymbol{u}^{f}(t, k)\right\|_{\lambda} \\
& \leq\left\|I-\boldsymbol{\beta} F_{u}\right\|\left\|\boldsymbol{\delta} \boldsymbol{u}^{f}(t, k-1)\right\|_{\lambda}+\boldsymbol{\varepsilon}_{1} \boldsymbol{\varepsilon}_{4} \sup _{t \in[0, T]} \varepsilon_{3}^{-\lambda t}  \tag{42}\\
& +\boldsymbol{\varepsilon}_{1} \sup _{t[0, T]} \varepsilon_{3}^{-\lambda t} \sum_{j=0}^{T-1}\left[\left(\varepsilon_{3}^{T-j}+\boldsymbol{\varepsilon}_{3}^{T-j-1}+\boldsymbol{\varepsilon}_{3}^{T-j-\varepsilon_{e, t}+1}\right)\left\|\boldsymbol{\delta} \boldsymbol{u}^{f}(j, k-1)\right\|\right]
\end{align*}
where,
\begin{align*}
& \sup _{t \in[0, T]} \boldsymbol{\varepsilon}_{3}^{-\lambda t} \sum_{j=0}^{T-1}\left[\left(\varepsilon_{3}^{T-j}+\boldsymbol{\varepsilon}_{3}^{T-j-1}+\boldsymbol{\varepsilon}_{3}^{T-j-l_{e, t}+1}\right)\left\|\boldsymbol{\delta} \mathbf{u}^{f}(j, k-1)\right\|\right] \\
& \leq \sup _{t \in[0, T]} \sum_{j=0}^{T-1}\left[\left(\sup _{t \in[0, T]} \boldsymbol{\varepsilon}_{3}^{-\lambda t}\left\|\boldsymbol{\delta} \boldsymbol{u}^{f}(j, k-1)\right\|\right) \|\left(\boldsymbol{\varepsilon}_{3}^{T-j}+\boldsymbol{\varepsilon}_{3}^{T-j-1}+\boldsymbol{\varepsilon}_{3}^{T-j-l_{e, t}, 1}\right)^{(1-\lambda)}\right] \\
& \leq\left\|\boldsymbol{\delta} \mathbf{u}^{f}(t, k-1)\right\|_{\lambda} \sup _{t \in[0, T]} \sum_{j=0}^{T-1}\left(\boldsymbol{\varepsilon}_{5}^{(T-j)}\right)^{(1-\lambda)}  \tag{43}\\
& =\left\|\boldsymbol{\delta} \mathbf{u}^{f}(t, k-1)\right\|_{\lambda} \frac{1-\boldsymbol{\varepsilon}_{5}^{-(\lambda-1) T}}{\boldsymbol{\varepsilon}_{5}^{\lambda-1}-1}
\end{align*}
where, $\varepsilon_{5}=4 \max _{t \in[0, T]}\left\{\varepsilon_{3}, 1,2 \varepsilon_{3}^{-1}, \varepsilon_{3}^{l_{e, t}-2}\right\} \geq 1$.\\

Substituting formula (43) into (42) yields:
\begin{align*}
& \left\|\boldsymbol{\delta} \mathbf{u}^{f}(t, k)\right\|_{\lambda} \\
& \leq\left\|I-\boldsymbol{\beta} F_{u}\right\|\left\|\boldsymbol{\delta} \mathbf{u}^{f}(t, k-1)\right\|_{\lambda}+\varepsilon_{1} \varepsilon_{4} \sup _{t \in[0, T]} \varepsilon_{3}^{-\lambda t}+\varepsilon_{1}\left\|\boldsymbol{\delta} \mathbf{u}^{f}(t, k-1)\right\|_{\lambda} \frac{1-\boldsymbol{\varepsilon}_{5}^{-(\lambda-1) T}}{\varepsilon_{5}^{\lambda-1}-1}  \tag{44}\\
& \leq\left(\left\|I-\boldsymbol{\beta} F_{u}\right\|+\varepsilon_{1} \frac{1-\boldsymbol{\varepsilon}_{5}^{-(\lambda-1) T}}{\varepsilon_{5}^{\lambda-1}-1}\right)\left\|\delta \boldsymbol{u}^{f}(t, k-1)\right\|_{\lambda}+\varepsilon_{1} \varepsilon_{4} \sup _{t \in[0, T]} \varepsilon_{3}^{-\lambda t}
\end{align*}

From formula (44), we can see that there exists a sufficiently large $\boldsymbol{\lambda}$ such that:
\begin{align*}
&\left\|I-\boldsymbol{\beta} F_{u}\right\|+\varepsilon_{1} \frac{1-\boldsymbol{\varepsilon}_{5}^{-(\lambda-1) T}}{\varepsilon_{5}^{\lambda-1}-1}=\boldsymbol{\rho}<1  \tag{45}\\
&\left\|\delta \mathbf{u}^{f}(t, k)\right\|_{\lambda} \leq \boldsymbol{\rho}\left\|\boldsymbol{\delta} \boldsymbol{u}^{f}(t, k)\right\|_{\lambda}+\varepsilon_{1} \varepsilon_{4} \\
& \leq \boldsymbol{\rho}^{k}\left\|\boldsymbol{\delta} \boldsymbol{u}^{f}(t, k)\right\|_{\lambda}+\varepsilon_{1} \varepsilon_{4} \frac{1-\boldsymbol{\rho}^{k}}{1-\boldsymbol{\rho}} \tag{46}
\end{align*}

when $k \rightarrow \infty$, have :
\begin{equation*}
\lim _{k \rightarrow \infty}\left\|\delta u^{f}(t, k)\right\|_{\lambda} \leq \frac{\varepsilon_{1} \varepsilon_{4}}{1-\rho} \tag{47}
\end{equation*}

According to formula (39), we can get:
\begin{align*}
& \lim _{k \rightarrow \infty}\left(\left\|\boldsymbol{u}^{b}(t, k-1)\right\|_{\lambda}+\|\boldsymbol{e}(t, k-1)\|_{\lambda}\right) \\
& \leq\left(\left\|\boldsymbol{u}^{b}(0, k-1)\right\|+\|\boldsymbol{e}(0, k-1)\|\right)  \tag{48}\\
& +\frac{1-\boldsymbol{\varepsilon}_{5}^{-(\lambda-1) T}}{\boldsymbol{\varepsilon}_{5}^{\lambda-1}-1} \lim _{k \rightarrow \infty}\left\|\boldsymbol{\delta} \boldsymbol{u}^{f}(t, k-1)\right\|_{\lambda}
\end{align*}

According to formulas (47) and (48), $\lim _{k \rightarrow \infty}\left\|\delta \boldsymbol{u}^{f}(t, k)\right\| \leq \sigma, \lim _{k \rightarrow \infty}\|\boldsymbol{e}(t, k)\| \leq \boldsymbol{\sigma}$, and $\sigma=\frac{\boldsymbol{\varepsilon}_{1} \varepsilon_{4}}{1-\boldsymbol{\rho}}>0$ are a constant. When condition $\|\boldsymbol{e}(0, k)\|=0$ and $\left\|\boldsymbol{u}^{b}(0, k)\right\|=0$ are met, then we get condition $\boldsymbol{\varepsilon}_{4}=0$, and then we get condition $\lim _{k \rightarrow \infty}\left\|\delta \boldsymbol{u}^{f}(t, k)\right\|_{\lambda}=0, \lim _{k \rightarrow \infty}\left\|\boldsymbol{u}^{b}(t, k)\right\|_{\lambda}=0$, and, $\lim _{k \rightarrow \infty}\|\boldsymbol{e}(t, k)\|_{\lambda}=0$. So far, the proof is complete.
\end{appendix}

\end{document}